\documentclass[10pt,twocolumn,letterpaper]{article}

\usepackage{times}
\usepackage{microtype} 
\usepackage{graphicx} 
\usepackage{subfigure} 

\usepackage{algorithm}
\usepackage{algorithmic}

\renewcommand{\algorithmicrequire}{\textbf{Initialize:}}   
\renewcommand{\algorithmicensure}{\textbf{Iteration:}} 
\renewcommand{\algorithmicinput}{\textbf{Input:}} 
\renewcommand{\algorithmicoutput}{\textbf{Output:}}

\usepackage{helvet}
\usepackage{courier}

\usepackage{makecell}
\usepackage{graphicx}
\usepackage{subfigure}
\usepackage{color}
\usepackage{amsfonts}
\usepackage{amsmath}
\usepackage{amssymb}

\usepackage{multirow}
\usepackage{booktabs}

\def\grad{\mbox{\text{grad}}}

\def\tr{\mbox{tr}}

\def\ie{\mbox{\textit{i.e.} }}
\def\eg{\mbox{\textit{e.g.} }}
\def\wrt{\mbox{\textit{w.r.t. }}}

\def\bgamma{\mbox{{\boldmath $\gamma$}}}

\def\bmu{\mbox{{\boldmath $\mu$}}}

\def\bxi{\mbox{{\boldmath $\xi$}}}

\def\bSigma{\mbox{{\boldmath $\Sigma$}}}

\def\mA{{\mathcal A}}
\def\mB{{\mathcal B}}
\def\mC{{\mathcal C}}
\def\mD{{\mathcal D}}
\def\mE{{\mathcal E}}
\def\mF{{\mathcal F}}
\def\mG{{\mathcal G}}
\def\mH{{\mathcal H}}
\def\mI{{\mathcal I}}
\def\mJ{{\mathcal J}}
\def\mK{{\mathcal K}}
\def\mL{{\mathcal L}}
\def\mM{{\mathcal M}}
\def\mN{{\mathcal N}}
\def\mO{{\mathcal O}}
\def\mP{{\mathcal P}}
\def\mQ{{\mathcal Q}}
\def\mR{{\mathcal R}}
\def\mS{{\mathcal S}}
\def\mT{{\mathcal T}}
\def\mU{{\mathcal U}}
\def\mV{{\mathcal V}}
\def\mW{{\mathcal W}}
\def\mX{{\mathcal X}}
\def\mY{{\mathcal Y}}
\def\mZ{{\mathcal{Z}}}

\DeclareMathAlphabet\mathbfcal{OMS}{cmsy}{b}{n}

\def\0{{\bf 0}}
\def\1{{\bf 1}}

\def\ba{{\bf a}}
\def\bb{{\bf b}}
\def\bc{{\bf c}}
\def\bd{{\bf d}}
\def\be{{\bf e}}
\def\bff{{\bf f}}
\def\bg{{\bf g}}
\def\bh{{\bf h}}

\def\bj{{\bf j}}
\def\bk{{\bf k}}
\def\bl{{\bf l}}
\def\bn{{\bf n}}
\def\bo{{\bf o}}
\def\bp{{\bf p}}
\def\bq{{\bf q}}
\def\br{{\bf r}}
\def\bs{{\bf s}}
\def\bt{{\bf t}}
\def\bu{{\bf u}}
\def\bv{{\bf v}}
\def\bw{{\bf w}}
\def\bx{{\bf x}}
\def\by{{\bf y}}
\def\bz{{\bf z}}

\def\mmE{{\mathbb E}}

\def\tC{\tilde{C}}

\def\tJ{\tilde{J}}

\def\tL{\tilde{L}}

\def\pd{{\succ\0}}
\def\psd{{\succeq\0}}
\def\vphi{\varphi}
\def\trsp{{\sf T}}

\def\st{{\mathrm{s.t.}}}

\def\bx{{\bf x}}

\def\by{{\bf y}}

\def\bw{{\bf w}}

\def\bmu{{\bm \mu}}

\def\bb{{\bf b}}
\def\bg{{\bf g}}
\def\bp{{\bf p}}

\def\bh{{\bf h}}
\def\bc{{\bf c}}
\def\bz{{\bf z}}

\def\st{{\mathrm{s.t.}}}
\def\tr{\mathrm{tr}}
\def\grad{{\mathrm{grad}}}

\usepackage{ntheorem}

\newtheorem{thm}{Theorem}
\newtheorem*{*thm}{Theorem}

\newtheorem*{*lemma}{Lemma}

\newenvironment*{proof}{\textbf{Proof}\quad}{\hfill $\square$\par}

\usepackage{enumerate}
\usepackage{enumitem}

\def\mid{\text{mid}}

\def\red{\textcolor{red}}
\def\blue{\textcolor{blue}}

\usepackage{iccv}
\usepackage{times}
\usepackage{epsfig}
\usepackage{graphicx}
\usepackage{amsmath}
\usepackage{amssymb}
\usepackage{balance}

\usepackage[pagebackref=true,breaklinks=true,letterpaper=true,colorlinks,bookmarks=false]{hyperref}

\iccvfinalcopy 

\def\iccvPaperID{1134} 

\ificcvfinal\fi
\begin{document}



\newcommand{\figleft}{{\em (Left)}}
\newcommand{\figcenter}{{\em (Center)}}
\newcommand{\figright}{{\em (Right)}}
\newcommand{\figtop}{{\em (Top)}}
\newcommand{\figbottom}{{\em (Bottom)}}
\newcommand{\captiona}{{\em (a)}}
\newcommand{\captionb}{{\em (b)}}
\newcommand{\captionc}{{\em (c)}}
\newcommand{\captiond}{{\em (d)}}

\newcommand{\newterm}[1]{{\bf #1}}

\def\figref#1{figure~\ref{#1}}
\def\Figref#1{Figure~\ref{#1}}
\def\twofigref#1#2{figures \ref{#1} and \ref{#2}}
\def\quadfigref#1#2#3#4{figures \ref{#1}, \ref{#2}, \ref{#3} and \ref{#4}}
\def\secref#1{section~\ref{#1}}
\def\Secref#1{Section~\ref{#1}}
\def\twosecrefs#1#2{sections \ref{#1} and \ref{#2}}
\def\secrefs#1#2#3{sections \ref{#1}, \ref{#2} and \ref{#3}}
\def\eqref#1{equation~\ref{#1}}
\def\Eqref#1{Equation~\ref{#1}}
\def\plaineqref#1{\ref{#1}}
\def\chapref#1{chapter~\ref{#1}}
\def\Chapref#1{Chapter~\ref{#1}}
\def\rangechapref#1#2{chapters\ref{#1}--\ref{#2}}
\def\algref#1{algorithm~\ref{#1}}
\def\Algref#1{Algorithm~\ref{#1}}
\def\twoalgref#1#2{algorithms \ref{#1} and \ref{#2}}
\def\Twoalgref#1#2{Algorithms \ref{#1} and \ref{#2}}
\def\partref#1{part~\ref{#1}}
\def\Partref#1{Part~\ref{#1}}
\def\twopartref#1#2{parts \ref{#1} and \ref{#2}}

\def\ceil#1{\lceil #1 \rceil}
\def\floor#1{\lfloor #1 \rfloor}
\def\1{\bm{1}}
\newcommand{\train}{\mathcal{D}}
\newcommand{\valid}{\mathcal{D_{\mathrm{valid}}}}
\newcommand{\test}{\mathcal{D_{\mathrm{test}}}}

\def\eps{{\epsilon}}

\def\reta{{\textnormal{$\eta$}}}
\def\ra{{\textnormal{a}}}
\def\rb{{\textnormal{b}}}
\def\rc{{\textnormal{c}}}
\def\rd{{\textnormal{d}}}
\def\re{{\textnormal{e}}}
\def\rf{{\textnormal{f}}}
\def\rg{{\textnormal{g}}}
\def\rh{{\textnormal{h}}}
\def\ri{{\textnormal{i}}}
\def\rj{{\textnormal{j}}}
\def\rk{{\textnormal{k}}}
\def\rl{{\textnormal{l}}}
\def\rn{{\textnormal{n}}}
\def\ro{{\textnormal{o}}}
\def\rp{{\textnormal{p}}}
\def\rq{{\textnormal{q}}}
\def\rr{{\textnormal{r}}}
\def\rs{{\textnormal{s}}}
\def\rt{{\textnormal{t}}}
\def\ru{{\textnormal{u}}}
\def\rv{{\textnormal{v}}}
\def\rw{{\textnormal{w}}}
\def\rx{{\textnormal{x}}}
\def\ry{{\textnormal{y}}}
\def\rz{{\textnormal{z}}}

\def\rvepsilon{{\mathbf{\epsilon}}}
\def\rvtheta{{\mathbf{\theta}}}
\def\rva{{\mathbf{a}}}
\def\rvb{{\mathbf{b}}}
\def\rvc{{\mathbf{c}}}
\def\rvd{{\mathbf{d}}}
\def\rve{{\mathbf{e}}}
\def\rvf{{\mathbf{f}}}
\def\rvg{{\mathbf{g}}}
\def\rvh{{\mathbf{h}}}
\def\rvu{{\mathbf{i}}}
\def\rvj{{\mathbf{j}}}
\def\rvk{{\mathbf{k}}}
\def\rvl{{\mathbf{l}}}
\def\rvm{{\mathbf{m}}}
\def\rvn{{\mathbf{n}}}
\def\rvo{{\mathbf{o}}}
\def\rvp{{\mathbf{p}}}
\def\rvq{{\mathbf{q}}}
\def\rvr{{\mathbf{r}}}
\def\rvs{{\mathbf{s}}}
\def\rvt{{\mathbf{t}}}
\def\rvu{{\mathbf{u}}}
\def\rvv{{\mathbf{v}}}
\def\rvw{{\mathbf{w}}}
\def\rvx{{\mathbf{x}}}
\def\rvy{{\mathbf{y}}}
\def\rvz{{\mathbf{z}}}

\def\ba{{\mathbf{a}}}
\def\bb{{\mathbf{b}}}
\def\bc{{\mathbf{c}}}
\def\bd{{\mathbf{d}}}
\def\be{{\mathbf{e}}}
\def\bff{{\mathbf{f}}}
\def\bg{{\mathbf{g}}}
\def\bh{{\mathbf{h}}}
\def\bu{{\mathbf{i}}}
\def\bj{{\mathbf{j}}}
\def\bk{{\mathbf{k}}}
\def\bl{{\mathbf{l}}}
\def\bm{{\mathbf{m}}}
\def\bn{{\mathbf{n}}}
\def\bo{{\mathbf{o}}}
\def\bp{{\mathbf{p}}}
\def\bq{{\mathbf{q}}}
\def\br{{\mathbf{r}}}
\def\bs{{\mathbf{s}}}
\def\bt{{\mathbf{t}}}
\def\bu{{\mathbf{u}}}
\def\bv{{\mathbf{v}}}
\def\bw{{\mathbf{w}}}
\def\bx{{\mathbf{x}}}
\def\by{{\mathbf{y}}}
\def\bz{{\mathbf{z}}}

\def\erva{{\textnormal{a}}}
\def\ervb{{\textnormal{b}}}
\def\ervc{{\textnormal{c}}}
\def\ervd{{\textnormal{d}}}
\def\erve{{\textnormal{e}}}
\def\ervf{{\textnormal{f}}}
\def\ervg{{\textnormal{g}}}
\def\ervh{{\textnormal{h}}}
\def\ervi{{\textnormal{i}}}
\def\ervj{{\textnormal{j}}}
\def\ervk{{\textnormal{k}}}
\def\ervl{{\textnormal{l}}}
\def\ervm{{\textnormal{m}}}
\def\ervn{{\textnormal{n}}}
\def\ervo{{\textnormal{o}}}
\def\ervp{{\textnormal{p}}}
\def\ervq{{\textnormal{q}}}
\def\ervr{{\textnormal{r}}}
\def\ervs{{\textnormal{s}}}
\def\ervt{{\textnormal{t}}}
\def\ervu{{\textnormal{u}}}
\def\ervv{{\textnormal{v}}}
\def\ervw{{\textnormal{w}}}
\def\ervx{{\textnormal{x}}}
\def\ervy{{\textnormal{y}}}
\def\ervz{{\textnormal{z}}}

\def\rmA{{\mathbf{A}}}
\def\rmB{{\mathbf{B}}}
\def\rmC{{\mathbf{C}}}
\def\rmD{{\mathbf{D}}}
\def\rmE{{\mathbf{E}}}
\def\rmF{{\mathbf{F}}}
\def\rmG{{\mathbf{G}}}
\def\rmH{{\mathbf{H}}}
\def\rmI{{\mathbf{I}}}
\def\rmJ{{\mathbf{J}}}
\def\rmK{{\mathbf{K}}}
\def\rmL{{\mathbf{L}}}
\def\rmM{{\mathbf{M}}}
\def\rmN{{\mathbf{N}}}
\def\rmO{{\mathbf{O}}}
\def\rmP{{\mathbf{P}}}
\def\rmQ{{\mathbf{Q}}}
\def\rmR{{\mathbf{R}}}
\def\rmS{{\mathbf{S}}}
\def\rmT{{\mathbf{T}}}
\def\rmU{{\mathbf{U}}}
\def\rmV{{\mathbf{V}}}
\def\rmW{{\mathbf{W}}}
\def\rmX{{\mathbf{X}}}
\def\rmY{{\mathbf{Y}}}
\def\rmZ{{\mathbf{Z}}}

\def\ermA{{\textnormal{A}}}
\def\ermB{{\textnormal{B}}}
\def\ermC{{\textnormal{C}}}
\def\ermD{{\textnormal{D}}}
\def\ermE{{\textnormal{E}}}
\def\ermF{{\textnormal{F}}}
\def\ermG{{\textnormal{G}}}
\def\ermH{{\textnormal{H}}}
\def\ermI{{\textnormal{I}}}
\def\ermJ{{\textnormal{J}}}
\def\ermK{{\textnormal{K}}}
\def\ermL{{\textnormal{L}}}
\def\ermM{{\textnormal{M}}}
\def\ermN{{\textnormal{N}}}
\def\ermO{{\textnormal{O}}}
\def\ermP{{\textnormal{P}}}
\def\ermQ{{\textnormal{Q}}}
\def\ermR{{\textnormal{R}}}
\def\ermS{{\textnormal{S}}}
\def\ermT{{\textnormal{T}}}
\def\ermU{{\textnormal{U}}}
\def\ermV{{\textnormal{V}}}
\def\ermW{{\textnormal{W}}}
\def\ermX{{\textnormal{X}}}
\def\ermY{{\textnormal{Y}}}
\def\ermZ{{\textnormal{Z}}}

\def\vzero{{\bm{0}}}
\def\vone{{\bm{1}}}
\def\vmu{{\bm{\mu}}}
\def\vtheta{{\bm{\theta}}}
\def\va{{\bm{a}}}
\def\vb{{\bm{b}}}
\def\vc{{\bm{c}}}
\def\vd{{\bm{d}}}
\def\ve{{\bm{e}}}
\def\vf{{\bm{f}}}
\def\vg{{\bm{g}}}
\def\vh{{\bm{h}}}
\def\vi{{\bm{i}}}
\def\vj{{\bm{j}}}
\def\vk{{\bm{k}}}
\def\vl{{\bm{l}}}
\def\vm{{\bm{m}}}
\def\vn{{\bm{n}}}
\def\vo{{\bm{o}}}
\def\vp{{\bm{p}}}
\def\vq{{\bm{q}}}
\def\vr{{\bm{r}}}
\def\vs{{\bm{s}}}
\def\vt{{\bm{t}}}
\def\vu{{\bm{u}}}
\def\vv{{\bm{v}}}
\def\vw{{\bm{w}}}
\def\vx{{\bm{x}}}
\def\vy{{\bm{y}}}
\def\vz{{\bm{z}}}

\def\evalpha{{\alpha}}
\def\evbeta{{\beta}}
\def\evepsilon{{\epsilon}}
\def\evlambda{{\lambda}}
\def\evomega{{\omega}}
\def\evmu{{\mu}}
\def\evpsi{{\psi}}
\def\evsigma{{\sigma}}
\def\evtheta{{\theta}}
\def\eva{{a}}
\def\evb{{b}}
\def\evc{{c}}
\def\evd{{d}}
\def\eve{{e}}
\def\evf{{f}}
\def\evg{{g}}
\def\evh{{h}}
\def\evi{{i}}
\def\evj{{j}}
\def\evk{{k}}
\def\evl{{l}}
\def\evm{{m}}
\def\evn{{n}}
\def\evo{{o}}
\def\evp{{p}}
\def\evq{{q}}
\def\evr{{r}}
\def\evs{{s}}
\def\evt{{t}}
\def\evu{{u}}
\def\evv{{v}}
\def\evw{{w}}
\def\evx{{x}}
\def\evy{{y}}
\def\evz{{z}}

\def\mA{{\bf{A}}}
\def\mB{{\bf{B}}}
\def\mC{{\bf{C}}}
\def\mD{{\bf{D}}}
\def\mE{{\bf{E}}}
\def\mF{{\bf{F}}}
\def\mG{{\bf{G}}}
\def\mH{{\bf{H}}}
\def\mI{{\bf{I}}}
\def\mJ{{\bf{J}}}
\def\mK{{\bf{K}}}
\def\mL{{\bf{L}}}
\def\mM{{\bf{M}}}
\def\mN{{\bf{N}}}
\def\mO{{\bf{O}}}
\def\mP{{\bf{P}}}
\def\mQ{{\bf{Q}}}
\def\mR{{\bf{R}}}
\def\mS{{\bf{S}}}
\def\mT{{\bf{T}}}
\def\mU{{\bf{U}}}
\def\mV{{\bf{V}}}
\def\mW{{\bf{W}}}
\def\mX{{\bf{X}}}
\def\mY{{\bf{Y}}}
\def\mZ{{\bf{Z}}}
\def\mBeta{{\bf{\beta}}}
\def\mPhi{{\bf{\Phi}}}
\def\mLambda{{\bf{\Lambda}}}
\def\mSigma{{\bf{\Sigma}}}
\def\bgamma{\mbox{{\boldmath $\gamma$}}}

\newcommand{\tens}[1]{\bm{\mathsfit{#1}}}
\def\tA{{\tens{A}}}
\def\tB{{\tens{B}}}
\def\tC{{\tens{C}}}
\def\tD{{\tens{D}}}
\def\tE{{\tens{E}}}
\def\tF{{\tens{F}}}
\def\tG{{\tens{G}}}
\def\tH{{\tens{H}}}
\def\tI{{\tens{I}}}
\def\tJ{{\tens{J}}}
\def\tK{{\tens{K}}}
\def\tL{{\tens{L}}}
\def\tM{{\tens{M}}}
\def\tN{{\tens{N}}}
\def\tO{{\tens{O}}}
\def\tP{{\tens{P}}}
\def\tQ{{\tens{Q}}}
\def\tR{{\tens{R}}}
\def\tS{{\tens{S}}}
\def\tT{{\tens{T}}}
\def\tU{{\tens{U}}}
\def\tV{{\tens{V}}}
\def\tW{{\tens{W}}}
\def\tX{{\tens{X}}}
\def\tY{{\tens{Y}}}
\def\tZ{{\tens{Z}}}

\def\gA{{\mathcal{A}}}
\def\gB{{\mathcal{B}}}
\def\gC{{\mathcal{C}}}
\def\gD{{\mathcal{D}}}
\def\gE{{\mathcal{E}}}
\def\gF{{\mathcal{F}}}
\def\gG{{\mathcal{G}}}
\def\gH{{\mathcal{H}}}
\def\gI{{\mathcal{I}}}
\def\gJ{{\mathcal{J}}}
\def\gK{{\mathcal{K}}}
\def\gL{{\mathcal{L}}}
\def\gM{{\mathcal{M}}}
\def\gN{{\mathcal{N}}}
\def\gO{{\mathcal{O}}}
\def\gP{{\mathcal{P}}}
\def\gQ{{\mathcal{Q}}}
\def\gR{{\mathcal{R}}}
\def\gS{{\mathcal{S}}}
\def\gT{{\mathcal{T}}}
\def\gU{{\mathcal{U}}}
\def\gV{{\mathcal{V}}}
\def\gW{{\mathcal{W}}}
\def\gX{{\mathcal{X}}}
\def\gY{{\mathcal{Y}}}
\def\gZ{{\mathcal{Z}}}

\def\sA{{\mathbb{A}}}
\def\sB{{\mathbb{B}}}
\def\sC{{\mathbb{C}}}
\def\sD{{\mathbb{D}}}
\def\sF{{\mathbb{F}}}
\def\sG{{\mathbb{G}}}
\def\sH{{\mathbb{H}}}
\def\sI{{\mathbb{I}}}
\def\sJ{{\mathbb{J}}}
\def\sK{{\mathbb{K}}}
\def\sL{{\mathbb{L}}}
\def\sM{{\mathbb{M}}}
\def\sN{{\mathbb{N}}}
\def\sO{{\mathbb{O}}}
\def\sP{{\mathbb{P}}}
\def\sQ{{\mathbb{Q}}}
\def\sR{{\mathbb{R}}}
\def\sS{{\mathbb{S}}}
\def\sT{{\mathbb{T}}}
\def\sU{{\mathbb{U}}}
\def\sV{{\mathbb{V}}}
\def\sW{{\mathbb{W}}}
\def\sX{{\mathbb{X}}}
\def\sY{{\mathbb{Y}}}
\def\sZ{{\mathbb{Z}}}

\def\emLambda{{\Lambda}}
\def\emA{{A}}
\def\emB{{B}}
\def\emC{{C}}
\def\emD{{D}}
\def\emE{{E}}
\def\emF{{F}}
\def\emG{{G}}
\def\emH{{H}}
\def\emI{{I}}
\def\emJ{{J}}
\def\emK{{K}}
\def\emL{{L}}
\def\emM{{M}}
\def\emN{{N}}
\def\emO{{O}}
\def\emP{{P}}
\def\emQ{{Q}}
\def\emR{{R}}
\def\emS{{S}}
\def\emT{{T}}
\def\emU{{U}}
\def\emV{{V}}
\def\emW{{W}}
\def\emX{{X}}
\def\emY{{Y}}
\def\emZ{{Z}}
\def\emSigma{{\Sigma}}

\newcommand{\etens}[1]{\mathsfit{#1}}
\def\etLambda{{\etens{\Lambda}}}
\def\etA{{\etens{A}}}
\def\etB{{\etens{B}}}
\def\etC{{\etens{C}}}
\def\etD{{\etens{D}}}
\def\etE{{\etens{E}}}
\def\etF{{\etens{F}}}
\def\etG{{\etens{G}}}
\def\etH{{\etens{H}}}
\def\etI{{\etens{I}}}
\def\etJ{{\etens{J}}}
\def\etK{{\etens{K}}}
\def\etL{{\etens{L}}}
\def\etM{{\etens{M}}}
\def\etN{{\etens{N}}}
\def\etO{{\etens{O}}}
\def\etP{{\etens{P}}}
\def\etQ{{\etens{Q}}}
\def\etR{{\etens{R}}}
\def\etS{{\etens{S}}}
\def\etT{{\etens{T}}}
\def\etU{{\etens{U}}}
\def\etV{{\etens{V}}}
\def\etW{{\etens{W}}}
\def\etX{{\etens{X}}}
\def\etY{{\etens{Y}}}
\def\etZ{{\etens{Z}}}

\newcommand{\pdata}{p_{\rm{data}}}
\newcommand{\ptrain}{\hat{p}_{\rm{data}}}
\newcommand{\Ptrain}{\hat{P}_{\rm{data}}}
\newcommand{\pmodel}{p_{\rm{model}}}
\newcommand{\Pmodel}{P_{\rm{model}}}
\newcommand{\ptildemodel}{\tilde{p}_{\rm{model}}}
\newcommand{\pencode}{p_{\rm{encoder}}}
\newcommand{\pdecode}{p_{\rm{decoder}}}
\newcommand{\precons}{p_{\rm{reconstruct}}}

\newcommand{\laplace}{\mathrm{Laplace}} 

\newcommand{\E}{\mathbb{E}}
\newcommand{\Ls}{\mathcal{L}}
\newcommand{\R}{\mathbb{R}}
\newcommand{\emp}{\tilde{p}}
\newcommand{\lr}{\alpha}
\newcommand{\reg}{\lambda}
\newcommand{\rect}{\mathrm{rectifier}}
\newcommand{\softmax}{\mathrm{softmax}}
\newcommand{\sigmoid}{\sigma}
\newcommand{\softplus}{\zeta}
\newcommand{\KL}{{\mathrm{KL}}}
\newcommand{\MMD}{{\mathrm{MMD}}}
\newcommand{\GAN}{{\mathrm{GAN}}}
\newcommand{\WGAN}{{\mathrm{WGAN}}}
\newcommand{\VAE}{{\mathrm{VAE}}}
\newcommand{\Var}{\mathrm{Var}}
\newcommand{\standarderror}{\mathrm{SE}}
\newcommand{\Cov}{\mathrm{Cov}}
\newcommand{\normlzero}{L^0}
\newcommand{\normlone}{L^1}
\newcommand{\normltwo}{L^2}
\newcommand{\normlp}{L^p}
\newcommand{\normmax}{L^\infty}

\newcommand{\parents}{Pa} 

\let\ab\allowbreak


\def\1{{\mathds{1}}}

\def\tprimal{\text{primal}}
\def\primal{\emph{\text{primal}}}
\def\tdual{\text{dual}}
\def\dual{\emph{\text{dual}}}

\def\st{{\mathrm{s.t.}}}
\def\tr{\mathrm{tr}}
\def\grad{{\mathrm{grad}}}


\def\pd{{\succ\0}}
\def\psd{{\succeq\0}}
\def\vphi{\varphi}
\def\trsp{{\sf T}}

\def\bgamma{\mbox{{\boldmath $\gamma$}}}
\def\bmu{\mbox{{\boldmath $\mu$}}}
\def\bxi{\mbox{{\boldmath $\xi$}}}
\def\bpi{\mbox{{\boldmath $\pi$}}}
\def\bSigma{\mbox{{\boldmath $\Sigma$}}}

\def\defeq{{\mathop{\rm{=}}\limits^{\text{def}}}}

\newcommand{\Perp}{\protect\mathpalette{\protect\independenT}{\perp}}
\def\independenT#1#2{\mathrel{\rlap{$#1#2$}\mkern2mu{#1#2}}}

\def\red{\textcolor{red}}
\def\blue{\textcolor{blue}}

\definecolor{mypink}{rgb}{0.858, 0.188, 0.478}

\def\cao{\textcolor{black}}
\def\kui{\textcolor{black}}
\def\guo{\textcolor{black}}
\def\mo{\textcolor{black}}
\def\ch{\textcolor{black}}
\def\warming{\textcolor{black}}
\def\temp{\textcolor{black}}

\title{Joint Wasserstein Distribution Matching}

\author{
JieZhang Cao$^{1,2}$\thanks{Equal contributions.}, 
Langyuan Mo$^{1,2*}$, 
Qing Du$^{1*}$, 
Yong Guo$^1$, 
Peilin Zhao$^3$, 
Junzhou Huang$^3$, 
Mingkui Tan$^{1,2}$\thanks{Corresponding author}\\
$^1$School of Software Engineering, South China University of Technology, China\\
$^2$Guangzhou Laboratory, Guangzhou, China\\
$^3$Tencent AI Lab \\
{\tt\small \{secaojiezhang, selymo\}@mail.scut.edu.cn, guoyongcs@gmail.com, peilinzhao@hotmail.com} \\
{\tt\small jzhuang@uta.edu, \{mingkuitan, duqing\}@scut.edu.cn }	
}

\maketitle

\begin{abstract} 
	Joint distribution matching (JDM) problem,  which aims to learn bidirectional mappings to match joint distributions of two domains, occurs in many machine learning and computer vision applications. This problem, however, is very difficult due to two critical challenges: (i) it is often difficult to exploit sufficient information from the joint distribution to conduct the matching; (ii) this problem is hard to formulate and optimize. In this paper, relying on optimal transport theory, we propose to address JDM problem by minimizing the Wasserstein distance of the joint distributions in two domains. However, the resultant optimization problem is still intractable. We then propose an important theorem to reduce the intractable problem into a simple optimization problem, and  develop a novel method (called Joint Wasserstein Distribution Matching (JWDM)) to solve it. In the experiments, we apply our method to {unsupervised image translation} and {cross-domain video synthesis}. Both qualitative and quantitative comparisons demonstrate the superior performance of our method over several state-of-the-arts. 
\end{abstract}

\section{Introduction} \label{sec:introduction}
Joint distribution matching (JDM) seeks to learn the bidirectional mappings to match the joint distributions of unpaired data in two different domains. 
Note that this problem has many applications in computer vision, such as image translation~\cite{zhu2017unpaired, liu2017unsupervised} and video synthesis \cite{bashkirova2018unsupervised, wang2018video}.
Compared to the learning of marginal distribution in each individual domain, learning the joint distribution of two domains is more difficult and has the following two challenges.

The first key challenge, from a probabilistic modeling perspective, is {how} to exploit the joint distribution of unpaired {data} by learning the bidirectional mappings between two different domains.
In the unsupervised learning setting, there are two sets of {samples} drawn separately from two marginal distributions in two domains.
Based on the coupling theory \cite{lindvall2002lectures}, there exist an infinite set of joint distributions given two marginal distributions, and thus infinite bidirectional mappings between two different domains \kui{may exist}.
Therefore, directly learning the joint distribution without additional information between the marginal distributions is a highly ill-posed problem.
Recently, many studies~\cite{zhu2017unpaired, yi2017dualgan, kim2017learning} {have been proposed to} learn {the} mappings in two domains separately, which cannot learn cross-domain correlations. 
Therefore, how to exploit sufficient information from the joint distribution still remains an open question.

\cao{
The second critical challenge is how to formulate and optimize the joint distribution matching problem.
\temp{Most} existing methods~\cite{li2017alice, pmlr-v80-pu18a} do not directly measure the distance between joint distributions, which may result in the distribution mismatching issue.
To address this, one can directly apply some statistics divergence, \eg, Wasserstein distance, to measure the divergence of joint distributions.
However, the optimization may result in intractable computational cost and statistical difficulties~\cite{genevay2018learning}.
Therefore, it is important to design a new objective function and an effective optimization method for the joint distribution matching problem.
}

\cao{
In this paper, we propose a Joint Wasserstein Distribution Matching (JWDM) method. 
Specifically, for the first challenge, we use the optimal transport theory to exploit geometry information and correlations between different domains.
For the second challenge, we apply Wasserstein distance to measure the divergence between joint distributions in two domains, and optimize it based on an equivalence theorem.
}

\vspace{5pt}
The contributions are summarized as follows:
\vspace{-7pt}
\begin{itemize}[leftmargin=*]
	\item Relying on optimal transport theory, we propose a novel JWDM to solve the joint distribution matching problem. The proposed method is able to exploit sufficient information for JDM by learning correlations between domains instead of learning from individual domain.
	\item We derive an important theorem so that the intractable primal problem of minimizing Wasserstein distance between joint distributions can be \kui{readily} reduced to a simple optimization problem (see Theorem ~\ref{thm:equ_problem}).
	\item 
	We apply JWDM to unsupervised image translation and cross-domain video synthesis. JWDM obtains highly qualitative images and visually smooth videos in two domains.
	Experiments on real-world datasets show the superiority of the proposed method over several state-of-the-arts. 
\end{itemize}

\section{Related Work}\label{sec:related_work}
\noindent\textbf{Image-to-image translation.}
Recently, Generative adversarial networks (GAN)~\cite{goodfellow2014generative, pmlr-v80-cao18a, salimans2018improving, cao2019multi, guo2018dual}, 
Variational Auto-Encoders (VAE)~\cite{kingma2013auto} and Wasserstein Auto-Encoders (WAE)~\cite{tolstikhin2017wasserstein} have emerged as popular techniques for the image-to-image translation problem. 
\mo{Pix2pix~\cite{isola2017image} proposes a unified framework 
which has been extended to generate high-resolution images~\cite{wang2018pix2pixhd}.
Besides, recent studies also attempt to tackle the image-to-image problem on the unsupervised setting.}
CycleGAN~\cite{zhu2017unpaired}, DiscoGAN~\cite{kim2017learning} and DualGAN~\cite{yi2017dualgan}  
minimize the adversarial loss and the cycle-consistent loss in two domains \mo{separately, where the cycle-consistent constraint helps to learn cross-domain mappings.
SCANs~\cite{li2018scan} also adopts this constraint and proposes to use multi-stage translation to enable higher resolution image-to-image translation.
}

\mo{
To tackle the large shape deformation difficulty, Gokaslan \textit{et al.}~\cite{gokaslan2018deformation} designs a discriminator with dilated convolutions. 
More recently, HarmonicGAN~\cite{zhang2018harmonic} introduces a smoothness term to enforce consistent mappings during the translation.
Besides, Alami \textit{et al.}~\cite{alami2018attention} introduces unsupervised attention mechanisms to improve the translation quality.
Unfortunately, these works minimize the distribution divergence of images in two domains separately,
which may induce a joint distribution mismatching issue.}
{Other methods focus on learning joint distributions.} CoGAN~\cite{liu2016coupled} learns a joint distribution by enforcing a weight-sharing constraint.
\mo{However, it uses noise variable as input thus can not control the outputs, which limits it's effect of applications.}
Moreover, UNIT~\cite{liu2017unsupervised} builds upon CoGAN by using a shared-latent space assumption and the same weight-sharing constraint.

\vspace{5pt}
\noindent\textbf{Video synthesis.}
In this paper, we further consider cross-domain video synthesis problem. 
Since existing image-to-image methods~\cite{zhu2017unpaired, kim2017learning, yi2017dualgan} cannot be directly used in the video-to-video synthesis problem, we combine some video frame interpolation methods~\cite{zhou2016view, ji2017deep, niklaus2017video, liu2017video} to synthesize video.
\kui{Note that
	one may directly apply existing UNIT~\cite{liu2017unsupervised} technique to do cross-domain video synthesis, which, however, may result in significantly incoherent videos with low visual quality.}
Recently, a video-to-video translation method~\cite{bashkirova2018unsupervised}  translates videos from one domain to another domain, but it cannot conduct video frame interpolation.
Moreover, Vid2vid\cite{wang2018video} proposes a video-to-video synthesis method, but it cannot work for the unsupervised setting.

\section{Problem Definition}\label{subsec:pro_def} 
\noindent\textbf{Notations.}
\kui{We use calligraphic letters (\textit{e.g.}, $ \gX $) to denote the space, capital letters (\textit{e.g.}, $ X $) to denote random variables, and bold lower case letter (\textit{e.g.}, $ \rvx$) to be their corresponding values.}
We denote probability distributions with capital letters (\textit{i.e.}, $ P(X) $) and the corresponding densities with bold lower case letters (\textit{i.e.}, $ p(\rvx) $).
Let $ (\gX, P_X) $ be the domain, $ P_X $ be the marginal distribution over $ \gX $, and $ \gP(\gX) $ be the set of all the probability measures over $ \gX $.

\begin{figure*}[t]
	\centering
	{
		\includegraphics[width=1.8\columnwidth]{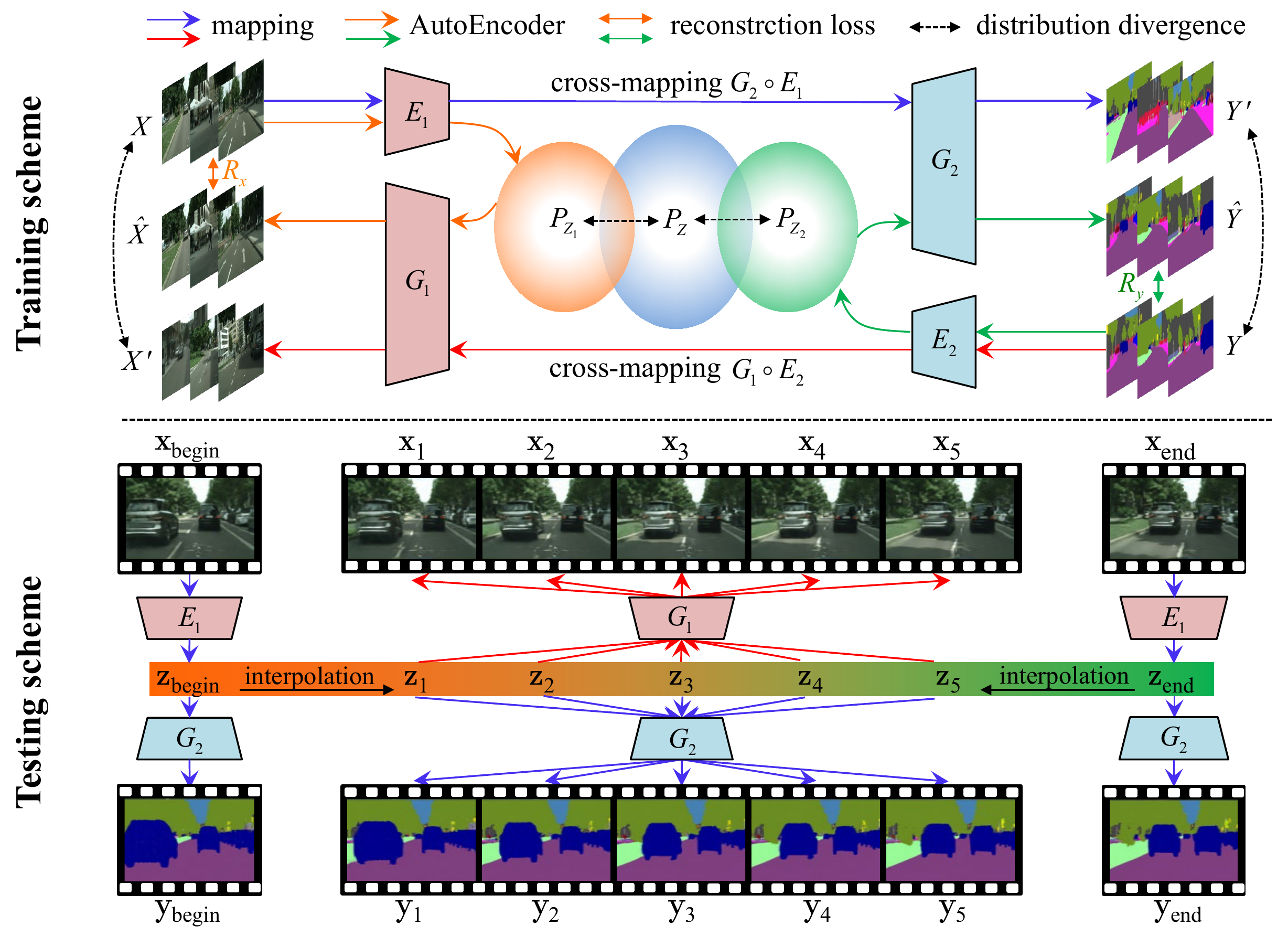}
		\caption{General training and testing scheme of Joint Wasserstein Distribution Matching for \textbf{unsupervised image translation} and \textbf{cross-domain video synthesis}.
		\textbf{Training scheme:} given real data $ X $ and $ Y $, we learn cross-domain mappings (\ie, $ G_2 \circ E_1 $ and $ G_1 \circ E_2 $) to generate samples $ Y' $ and $ X' $ such that the generated distributions can be close to real distribution. Moreover, latent distributions generated by Auto-Encoder (\ie, $ G_1 \circ E_1 $ and $ G_2 \circ E_2 $) should be close to each other.
  		\textbf{Testing scheme:} given two video frames (\ie, $\bx_{\text{begin}}$ and $\bx_{\text{end}}$) in the source domain, we extract their embeddings $\bz_{\text{begin}}$ and $\bz_{\text{end}}$, 
  		then we synthesize two videos in two domains by mapping linear interpolations to the corresponding frames.
  		More details for these two tasks can be found in Subsection \ref{subsec:app}.
		\label{fig:framework}
		}
	}
\end{figure*}


\vspace{5pt}
\noindent\textbf{Wasserstein distance.}
Recently, optimal transport \cite{villani2008optimal} has many applications \cite{arjovsky2017wasserstein, yan2019oversampling}. 
Based on optimal transport theory, we provide the definition of Wasserstein distance and then develop our proposed method. 
Given the distributions $P_X$ and $P_G$, where $P_G$ is generated by a generative model $G$, the Monge-Kantorovich problem is to find a transport plan (\textit{i.e.}, a joint distribution) $ P {\in} \gP(P_X, P_G) $ such that
\begin{align} \label{pro:WD}
\gW (P_X, P_G) {=} \inf_{P \in \gP(X {\sim} P_X, X' {\sim} P_G)} \E_{P} \left[ c(X, X') \right],
\end{align} 
where $ c(\cdot, \cdot) $ is a cost function and $ \gP(X {\sim} P_X, X' {\sim} P_G) $ is a set of all joint distributions with the marginals $ P_X $ and $ P_G $, respectively.
In next section, we use Wasserstein distance to measure the distribution divergence and develop our method.

\cao{
In this paper, we focus on the joint distribution matching problem which has broad applications like unsupervised image translation~\cite{zhu2017unpaired, liu2017unsupervised} and cross-domain video synthesis~\cite{wang2018video, bashkirova2018unsupervised}. 
For convenience, we first give a formal definition of joint distribution and the problem setting as follows.}

\vspace{10pt}
\noindent\textbf{Joint distribution matching problem.}
Let $ (\gX, P_X) $ and $ (\gY, P_Y) $ be two domains, where $ P_X $ and $ P_Y $ are the marginal distributions over $ \gX $ and $ \gY $, respectively.
In this paper, we seek to learn two cross-domain mappings $ f: \gX \to \gY $ and $ g: \gY \to \gX $ to construct two joint distributions defined as: 
\begin{align}
	P_{\gA} (X, Y') \quad \text{and}\quad P_{\gB} (X', Y),
\end{align}
where $ Y'{=}f(X) $ and $ X'{=}g(Y) $. 
In this paper, our goal is to match the joint distributions $ P_{\gA} $ and $ P_{\gB} $ as follows.

\vspace{10pt}
\noindent\textbf{Joint distribution matching problem.}
Relying on optimal transport theory, we employ Wasserstein distance (\ref{pro:WD}) to measure the distance between joint distributions $ P_{\gA} $ and $ P_{\gB} $, namely $ \gW(P_{\gA}, P_{\gB}) $.
To match these two joint distributions, we seek to learn the cross-domain mappings $f$ and $g$ by minimizing the following Wasserstein distance, \ie,
\begin{align} \label{problem:Wasserstein_jd}
\gW_c(P_{\gA},  P_{\gB}) {=} \min\limits_{P {\in} \gP(P_{\gA}, P_{\gB})}
\E_{P} [c(X, Y'; X', Y)],
\end{align}
where $ \gP(P_{\gA}, P_{\gB}) $ is the set of couplings composed of all joint distributions, 
and $ c $ is any measurable cost function. 
\cao{In the following, we propose to show how to solve Problem (\ref{problem:Wasserstein_jd}).}

\section{Joint Wasserstein Distribution Matching} \label{sec:JWDM}
\cao{In practice, directly optimizing Problem (\ref{problem:Wasserstein_jd}) would face two challenges.}
\textbf{First}, directly optimizing Problem (\ref{problem:Wasserstein_jd}) would incur intractable computational cost~\cite{genevay2018learning}. 
\textbf{Second}, how to choose an appropriate cost function is very difficult. 
To address these, we seek to reduce Problem (\ref{problem:Wasserstein_jd}) into a simpler optimization problem using the following theorem.

\begin{thm} \emph{(\textbf{Problem equivalence)}}\label{thm:equ_problem}
	Given two deterministic models $ P_{G_1} (X'|Z) $ and $ P_{G_2} (Y'|Z) $ as Dirac measures, \textit{i.e.},
	$ P_{G_1}({X'}|Z{=}\rvz){=}\delta_{G_1(\rvz)} $ and $ P_{G_2}(Y'|Z{=}\rvz){=}\delta_{G_2(\rvz)} $ for all $ \rvz {\in} \gZ $, we can rewrite Problem (\ref{problem:Wasserstein_jd}) as follows:
	\begin{align} \label{prob:equ_thm}
	\gW_c(P_{\gA}, P_{\gB})
	{=}& \inf_{Q  {\in} \gQ_1} \E_{P_{X}} \E_{Q{(Z_1|X)}}  [c_1(X, G_1(Z_1))] \\
	&\tiny{+}\inf_{Q {\in} \gQ_2} \E_{P_{Y}} \E_{Q{(Z_2|Y)}} [c_2(G_2(Z_2), Y)], \nonumber
	\end{align}
	where we define $ \gQ_1 {=} \{Q(Z_1|X) |\; Q {\in} \tilde{\gQ}, P_Y {=} Q_{Y} \} $ and $ \gQ_2 {=} \{Q(Z_2|Y) |\; Q {\in} \tilde{\gQ}, P_X {=} Q_{X} \} $ as the sets of all probabilistic encoders, respectively,  \cao{where $Q$ satisfies the set $\tilde{\gQ} {=} \{ Q| P_{Z_1} {=} Q_{Z_1}, P_{Z_2} {=} Q_{Z_2} \} $}.
\end{thm}
\vspace{-5pt}
\begin{proof}
	See supplementary materials for the proof.
\end{proof}

\vspace{10pt}
\noindent\textbf{Optimization problem.}
Based on Theorem \ref{thm:equ_problem}, we are able to optimize Wasserstein distance by optimizing the reconstruction losses of two auto-encoders when the generated joint distributions can match real joint distributions. 
Specifically, let  $ \gR_x (F)$ and $\gR_y(F) $ be the reconstruction losses of two auto-encoders ($ G_1 {\circ} E_1$ and $ G_2 {\circ} E_2 $), respectively, and let the models be $ F{=}\{E_1{,} E_2{,} G_1{,} G_2\} $, we optimize Problem (\ref{prob:equ_thm}) as follows:
\begin{equation}
\begin{aligned}  \label{prob:equ_problem}
	\min_{F}\gW_c(\gP_{\gA}, \gP_{\gB}) &= \gR_x(F) + \gR_y(F), \\
 \st \quad P_X &= Q_X, P_Y = Q_Y, P_Z = Q_Z,
\end{aligned}
\end{equation}
where $ P_X, P_Y $ and $ P_Z $ are real distributions, and $ Q_X, Q_Y $ and $ Q_Z $ are generated distributions. 
By minimizing Problem (\ref{prob:equ_problem}), the two reconstruction losses  $ \gR_x (F)$ and $\gR_y(F) $ will be minimized, meanwhile the generated distributions can match real distributions. 
The details of objective function and optimization are given below.

\vspace{10pt}
\noindent\textbf{Objective function.}
With the help of Theorem \ref{thm:equ_problem}, intractable Problem (\ref{problem:Wasserstein_jd}) can be turned into a simple optimization problem. 
To optimize Problem (\ref{prob:equ_problem}), we propose to enforce the constraints (\ie, $ P_X {=} Q_X, P_Y {=} Q_Y, P_Z {=} Q_Z $) by introducing distribution divergences to measure the distance between generated and real distribution.
Specifically, given two Auto-Encoders $ G_1 {\circ} E_1$ and $ G_2 {\circ} E_2 $, 
and let $ F{=}\{E_1{,} E_2{,} G_1{,} G_2\} $, we can instead optimize the following problem for joint distribution matching problem: 
\begin{align}\label{prob:new_obj}
	\min_{F}\gW_c(P_{\gA},& P_{\gB})
	= \gR_x(F) + \gR_y(F) \\
	+& \lambda_x d(P_X, Q_X) {+} \lambda_y d(P_Y, Q_Y) {+} \lambda_z d(P_Z, Q_Z), \nonumber
\end{align}
where $ d(\cdot, \cdot) $ is arbitrary distribution divergence between two distributions,
and $ \lambda_x, \lambda_y $ and $ \lambda_z $ are hyper-parameters.
Note that the proposed objective function involves two kinds of functions, namely the \textbf{reconstruction loss} (\ie, $ \gR_x $ and $ \gR_y $) and \textbf{distribution divergence} (\ie, $ d(P_X, Q_X), d(P_Y, Q_Y)$ and $ d(P_Z, Q_Z) $).
\kui{We depict each of them as follows.}

\vspace{10pt}
\noindent\textbf{(i) Reconstruction loss.}
To optimize Problem (\ref{prob:new_obj}), the reconstruction loss should be small, it means that the reconstruction of any input should be close to the input for the source and the target domain.
As shown in Figure \ref{fig:framework}, the reconstruction of $ \bx $ in the source domain can be derived from  Auto-Encoder (\ie, $ G_1(E_1(\rvx)) $) and cycle mapping (\ie, $ G_1(E_2(G_2(E_1(\rvx)))) $). 
Similarly, the reconstruction in the target domain can be learned in the same way.
Taking the source domain as an example, given an input $ \bx $, 
we minimize the following reconstruction loss
\begin{equation}
\begin{aligned}
\gR_x(F) =& \hat{\mmE}_{\bx \sim P_X} \left[ \left\| \rvx - G_1(E_1(\rvx)) \right\|_1 \right. \\
&{+} \left. \left\| \rvx - G_1(E_2(G_2(E_1(\rvx))))\right\|_1 \right] ,
\end{aligned}
\end{equation}
where $ \hat{\mmE}[\cdot] $ is the empirical expectation.
Note that the first term is the Auto-Encoders reconstruction loss and the second term is the cycle consistency loss \cite{zhu2017unpaired, liu2017unsupervised}.
For the target domain, the loss $ {\gR}_y(F)$ can be constructed similarly.

\begin{algorithm}[t]
	\caption{~Training details of JWDM.} 
	\label{Alg:JWDM}
	\begin{algorithmic}
		\INPUT Training data: $\{\bx_{i}\}_{i=1}^M$ and $\{\by_{j}\}_{j=1}^N$. \\ 
		\vspace{0.2 em}
		\OUTPUT Encoders $E_1, E_2$, decoders $ G_1, G_2$ and discriminators $D_x, D_y, D_z$.
		\vspace{0.2 em}
		\REQUIRE Models: $E_1, E_2, G_1, G_2, D_x, D_y, D_z$.
		\REPEAT
		\vspace{0.3 em}
		\STATE Update $ D_x, D_y, D_z $ by ascending:\\ 
		\vspace{0.4 em}
		~~~~~~~\small{$ \lambda_x d(P_X, Q_X) + \lambda_y d(P_Y, Q_Y) + \lambda_z d(P_Z, Q_Z) $} 
		\vspace{0.4 em} 
		\STATE Update $ E_1, E_2, G_1, G_2 $ by descending Loss (\ref{prob:new_obj})\\
		\UNTIL{models converged}
	\end{algorithmic}
\end{algorithm}

\vspace{10pt}
\noindent\textbf{(ii) Distribution divergence.}
From Theorem \ref{thm:equ_problem}, the constraints enforce that the generated distributions should be equal to the real distributions in the source and target domain.
Moreover, the latent distributions generated by two Auto-Encoders (\ie, $ G1\circ E1 $ and $ G2\circ E2 $) should be close to the prior distribution.
Therefore, there are three distribution divergence for the optimization, As shown in Figure \ref{fig:framework}.
Note that they are not limited to some specific distribution divergence, \eg, Adversarial loss (such as original GAN and WGAN \cite{arjovsky2017wasserstein}) and Maximum Mean Discrepancy (MMD), etc. 
In this paper, we use original GAN to measure the divergence between real and generated distribution. 
Taking $ d(P_X, Q_X) $ as an example, we minimize the following loss function, 
\begin{equation}
\begin{aligned}
d(P_X, Q_X)
= \max_{D_x} &\left[ \mathop{\hat\mmE}\nolimits_{\bx {\sim} {P}_X} \left[ \log D_x(\bx) \right] \right. \\
&+ \left. \mathop{\hat\mmE}\nolimits_{\tilde{\bx} {\sim} Q_X} \left[ \log (1-D_x(\tilde{\bx})) \right] \right],  
\end{aligned}
\end{equation}
where $ \bx $ denotes a sample drawn from real distribution $ P_X $, $ \tilde{\bx} $ denotes a sample drawn from generated distribution $ Q_X $, and $ D_x $ is a discriminator \wrt the source domain.
Similarly, the losses $d(P_Y, Q_Y) $ and $d(P_Z, Q_Z)$ can be constructed in the same way.
\footnote{Please find more details in supplementary materials.}  


\begin{algorithm}[t]
	\caption{{~Inference for cross-domain video synthesis.}}
	\label{Alg:Inference}
	\begin{algorithmic}
		\INPUT Testing data in the source domain: $\{\bx_{\text{begin}}, \bx_{\text{end}}\}$, \\
		the number of interpolation frames $n$.
		\STATE {Step 1: \textbf{Video frame interpolation}}\\
		~~~~~~~~~$ \bz_{\text{begin}} = E_1(\bx_{\text{begin}}),~~  \bz_{\text{end}} = E_1(\bx_{\text{end}})$\\
		~~~~~~~~~$ \bz_{\text{mid}} = \rho \bz_{\text{begin}} {\small +} (1 {\small -} \rho) \bz_{\text{end}}, ~~\rho {\in} (\frac{1}{n},...,\frac{n-1}{n}) $\\
		~~~~~~~~~$ \bx_{\text{mid}} = G_1(\bz_{\text{mid}}) $\\
		~~~~~~~~~Synthesized video: $\{ \bx_{\text{begin}}, \{\bx_{\text{mid}}\}^n_{\text{mid}=1}, \bx_{\text{end}} \}$
		\STATE Step 2: \textbf{Video translation}\\
		~~~~~~~~~$ \by_{\text{begin}} {=} G_2(\bz_{\text{begin}}), \by_{\text{mid}} {=} G_2(\bz_{\text{mid}}), \by_{\text{end}} {=} G_2(\bz_{\text{end}})$,\\  
		~~~~~~~~~Synthesized video: $\{ \by_{\text{begin}}, \{\by_{\text{mid}}\}^n_{\text{mid}}, \by_{\text{end}} \}$
	\end{algorithmic}
\end{algorithm}

\subsection{Connection to Applications} \label{subsec:app}
JWDM can be applied to unsupervised image translation and cross-domain video synthesis problems.

\vspace{10pt}
\noindent\textbf{(i) Unsupervised image translation.}
As shown in Figure \ref{fig:framework}, we can apply JWDM to solve unsupervised image translation problem. 
Specifically, given real data $ \bx $ and $ \by $ in the source and target domain, respectively, we learn cross-domain mappings (\ie, $ G_2 \circ E_1 $ and $ G_1 \circ E_2 $) to generate samples $ \by' $ and $ \bx' $.
By minimizing the distribution divergence $ {\gW}_c(P_{\gA}, P_{\gB})$, we learn cross-domain mappings 
\begin{align}
	\by'= G_2 \circ E_1(\bx) \quad \text{and} \quad \bx'= G_1 \circ E_2(\by),
\end{align}
where $ E_1 $ and $ E_2 $ are Encoders, and $ G_1 $ and $ G_2 $ are Decoders.
The detailed training method is shown in Algorithm~\ref{Alg:JWDM}.

\vspace{10pt}
\noindent\textbf{(ii) Cross-domain video synthesis.}
\kui{JWDM can be applied to conduct cross-domain video synthesis, which seeks to produce two videos in two different domains by performing linear interpolation based video synthesis~\cite{ji2017deep}.}
Specifically, given two input video frames $\bx_{\text{begin}}$ and $\bx_{\text{end}}$ in the source domain, we perform a linear interpolation between two embeddings extracted from $\bx_{\text{begin}}$ and $\bx_{\text{end}}$. Then the interpolated latent representations are decoded to the corresponding frames in the source and target domains (see Figure \ref{fig:framework} and Algorithm \ref{Alg:Inference}). The interpolated frames in two domains are 
\guo{
	\begin{equation}
	\begin{aligned}
	\bx_{\mid} {=} G_1 \left( \bz_{\mid} \right), \quad {\by}_{\mid} {=} G_2\left( \bz_{\mid} \right),
	\end{aligned}
	\end{equation}
	where $ \bz_{\mid} {=} \rho E_1(\bx_{\text{begin}}) {+} (1 {-} \rho) E_1(\bx_{\text{end}}), ~\rho \in (0,1) $, 
	$\bx_{\mid}$ denotes the interpolated frame in the source domain and $\by_{\mid}$ denotes the translated frame in the target domain. 
}

\section{Experiments}\label{sec:exp}
In this section, we first apply our proposed method on unsupervised image translation to evaluate the performance of joint distribution matching. Then, we further apply JWDM on cross-domain video synthesis to evaluate the interpolation performance on latent space.

\paragraph{Implementation details.}  
Adapted from CycleGAN~\cite{zhu2017unpaired}, the Auto-Encoders  (\ie $  G_1{\circ} E_1 $ and $ G_2{\circ} E_2 $) of JWDM are composed of two convolutional layers with the stride size of two for downsampling, 
six residual blocks, and two transposed convolutional layers with the stride size of two for upsampling.
We leverage PatchGANs~\cite{isola2017image} for the discriminator network.
\footnote{See supplementary materials for more details of network architectures.}
We follow the experimental settings in CycleGAN. For the optimization, we use Adam solver~\cite{kingma2014adam} with a mini-batch size of 1 to train the models, and use a learning rate of 0.0002 for the first 100 epochs and gradually decrease it to zero for the next 100 epochs. Following \cite{zhu2017unpaired}, we set \warming{$\lambda_{x} {=} \lambda_{y} {=} 0.1$} in Eqn.~(\ref{prob:new_obj}). 
By default, we set $\lambda_{z} {=} 0.1$ in our experiments (see more details in Subsection \ref{exp:influence}). 

\paragraph{Datasets.} 
We conduct experiments on two widely used benchmark datasets, \ie, Cityscapes~\cite{cordts2016cityscapes} and SYNTHIA~\cite{Ros_2016_CVPR}. 
\begin{itemize}
	\item \textbf{Cityscapes} \cite{cordts2016cityscapes} contains  $ 2048{\times} 1024$  street scene video of several German cities and a portion of ground truth semantic segmentation in the video.
\guo{In the experiment, we perform scene $\leftrightarrow$ segmentation translation in the unsupervised setting. } 
	\item\textbf{SYNTHIA} \cite{Ros_2016_CVPR}  
contains many synthetic videos in different scenes and seasons (\ie, spring, summer, fall and winter). 
In the experiment, we perform season translation from winter to the other three seasons. 
\end{itemize}

\paragraph{Evaluation metrics.}
For quantitative comparisons, we adopt Inception Score (IS), {Fr\'echet Inception Distance} (FID) and {video variant of FID} (FID4Video) to evaluate the generated samples.
\begin{itemize}
	\item \textbf{ Inception Score} (IS) \cite{salimans2016improved} is a widely used metric for generative models. By using the class predition information of Inception-V3 \cite{szegedy2016rethinking}, IS can be used to evaluate the quality and diversity of the generated samples.
	\item \textbf{Fr\'echet Inception Distance} (FID) \cite{heusel2017gans}  is another widely used metric for generative models. FID can evaluate the quality of the generated images because it captures the similarity of the generated samples to real ones and correlates well with human judgements. 
	\item \textbf{Video variant of FID} (FID4Video) \cite{wang2018video} evaluates both visual quality and temporal consistency of synthesized videos. Specifically, we use a pre-trained video recognition CNN (\ie I3D \cite{carreira2017quo}) as a feature extractor. Then, we use CNN to extract a spatio-temporal feature map for each video. Last, we calculate FID4Video using the formulation in \cite{wang2018video}.
\end{itemize}
In general, the higher IS means the better quality of translated images or videos. For both FID and FID4Video, the lower score means the better quality of translated images or videos.

\begin{figure*}[t]
	\centering
	\includegraphics[width = 1.75\columnwidth]{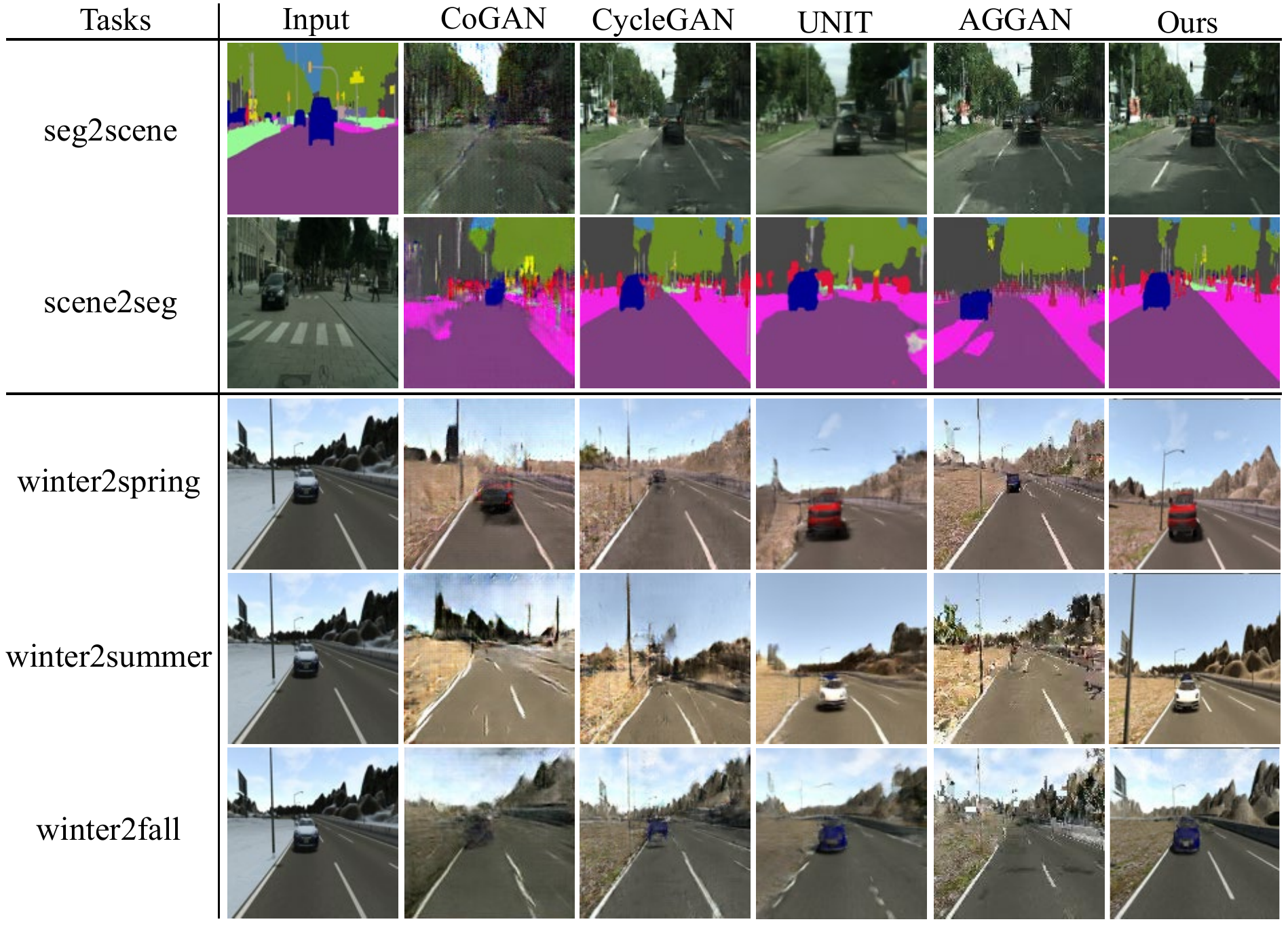}
	\caption{Comparisons with different methods for unsupervised image-to-image translation on Cityscape and SYNTHIA. }
	\label{fig:i2iresults}
\end{figure*}

\begin{table*}[t]
	\centering
	\caption{IS and FID scores of different methods for unsupervised image-to-image translation on Cityscape and SYNTHIA.}
	\label{tab:image2image}
	\resizebox{1\textwidth}{!}{
		\begin{tabular}{c|cc|cc|cc|cc|cc}
			\hline
			\multirow{2}{*}{method} & \multicolumn{2}{c|}{scene2segmentation} & \multicolumn{2}{c|}{segmentation2scene} & \multicolumn{2}{c|}{winter2spring} & \multicolumn{2}{c|}{winter2summer} & \multicolumn{2}{c}{winter2fall} \\
			\cline{2-11}
			& IS & FID & IS & FID & IS & FID & IS & FID & IS & FID \\
			\hline
			CoGAN~\cite{liu2016coupled}& 1.76 & 230.47 & 1.41 & 334.61 & 2.13 & 314.63 & 2.05 & 372.82 & 2.28 & 300.47 \\	
			CycleGAN~\cite{zhu2017unpaired}  & 1.83 & 87.69 & 1.70 & 124.49 & 2.23 & 115.43 & 2.32 & 120.21 & 2.30 & 100.30  \\
			UNIT~\cite{liu2017unsupervised} & 2.01 & 65.89 & 1.66 & 89.79 & 2.55 & 88.26 & 2.41 & 89.92 & 2.46 & 85.26 \\
			AGGAN~\cite{alami2018attention}  & 1.90 & 126.27 & 1.66 & 115.87 & 2.12 & 140.97 & 2.02 & 152.02 & 2.32 & 124.38 \\
			\hline
			JWDM (ours) & \textbf{2.42} & \textbf{21.89} & \textbf{1.92} & \textbf{42.13} & \textbf{3.12} & \textbf{82.34} & \textbf{2.86} & \textbf{84.37} & \textbf{2.85} & \textbf{83.26} \\
			\hline
		\end{tabular}%
	}
\end{table*}

\begin{figure*}[t]
	\centering
	\includegraphics[width = 2.1\columnwidth]{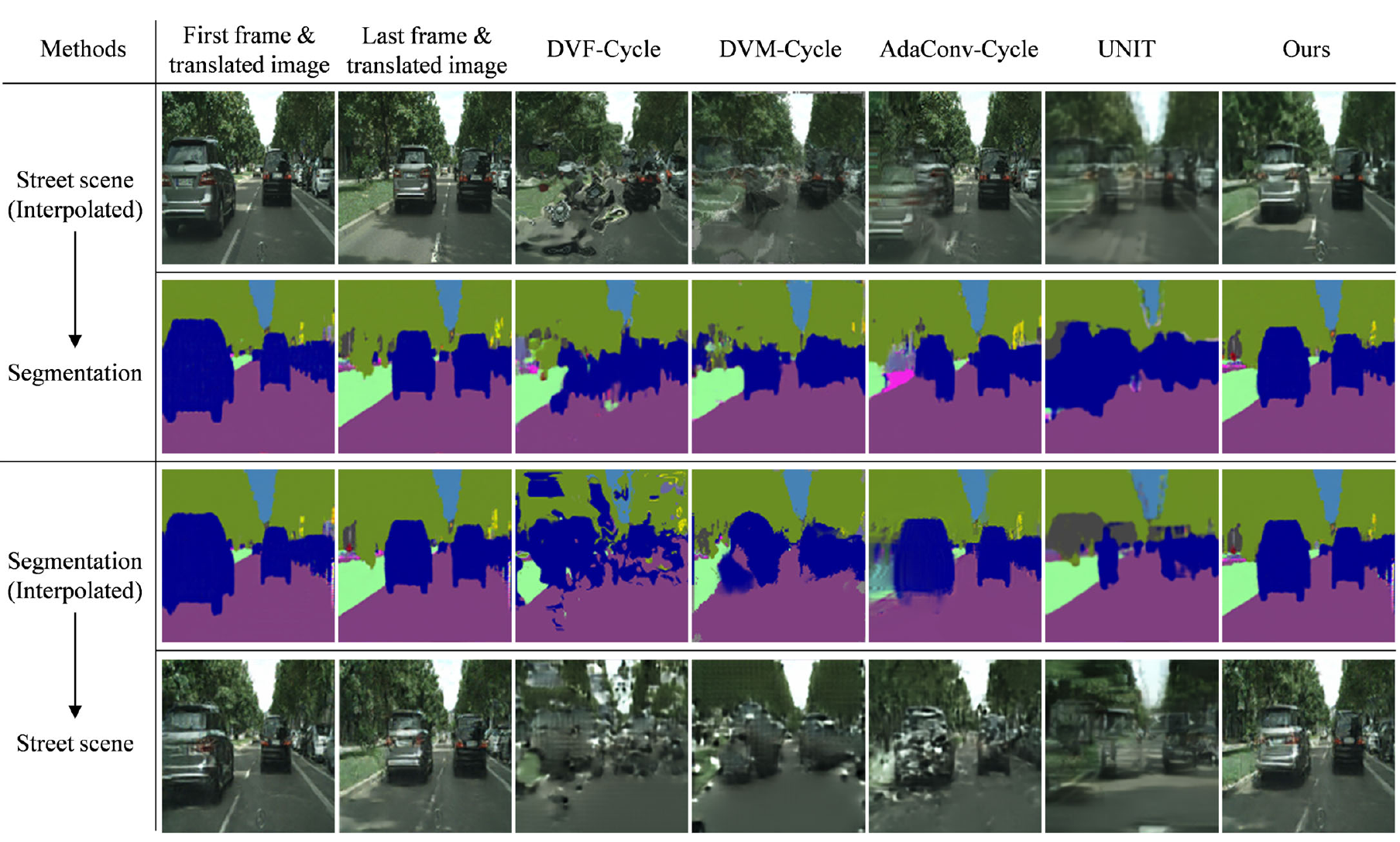}
	\caption{Comparisons of different methods for scene $\leftrightarrow$ segmentation translation on Cityscapes. We first synthesize a video of street scene and then translate it to the segmentation domain (Top), and vice versa for the mapping from segmentation to street scene (Bottom). 
	} 
	\label{fig:cityscapes1} 
\end{figure*}

\begin{table*}[tbp]
	\caption{Comparisons of different methods for cross-domain image-to-video synthesis on Cityscapes and SYNTHIA.}
	\label{tab:cityscapes_synthia}
	\centering
	\resizebox{1\textwidth}{!}{
		\begin{tabular}{c|ccc|ccc|ccc|ccc|ccc}
			\hline
			\multirow{2}[0]{*}{Method}
			& \multicolumn{3}{c|}{scene2segmentation} & \multicolumn{3}{c|}{segmentation2scene} & \multicolumn{3}{c|}{winter2spring} & \multicolumn{3}{c|}{winter2summer} & \multicolumn{3}{c}{winter2fall} \\
			\cline{2-16}
			& IS & FID & FID4Video & IS & FID & FID4Video & IS & FID & FID4Video & IS & FID & FID4Video & IS & FID & FID4Video \\
			\hline
			DVF-Cycle~\cite{liu2017video}& 1.43 & 110.59 & 23.95 & 1.34 & 151.27 & 40.61 & 2.09 & 152.44 & 42.22 & 2.07 & 160.69 & 42.43 & 2.44 & 163.13 & 41.04 \\	
			DVM-Cycle~\cite{ji2017deep}& 1.36 & 50.51 & 17.33 & 1.26 & 116.62 & 40.83 & 1.98 & 129.80 & 38.19 & 1.99 & 140.86 & 36.66 & 2.19 & 129.02 & 36.64  \\
			AdaConv-Cycle~\cite{niklaus2017video} & 1.29 & 33.50 & 14.96 & 1.27 & 99.67 & 30.24 & 1.91 & 117.40 & 23.83 & 2.10 &126.01 & 20.62 & 2.18 & 110.52 & 16.77 \\
			UNIT~\cite{liu2017unsupervised}& 1.66 & 31.27 & 10.12 & 1.89 & 76.72 & 29.21 & 2.14 & 96.40 & 23.12 & 2.13 & 108.01 & 24.70 & 2.30 & 97.73 & 20.39 \\
			Slomo-Cycle~\cite{jiang2018super} & \textbf{1.84} & 27.35 & 8.71 & 1.89 & 59.21 & 27.87 & 2.27 & 93.77 & 21.53 & 2.41 & 96.27 & 20.19 & 2.36 & 94.41 & 15.65 \\
			\hline
			JWDM (ours)   & 1.69 & \textbf{22.74} & \textbf{6.80} & \textbf{1.97} & \textbf{43.48} & \textbf{25.87} & \textbf{2.36} & \textbf{88.24} & \textbf{21.37} & \textbf{2.46} & \textbf{77.12} & \textbf{17.99} & \textbf{2.50} & \textbf{87.50} & \textbf{14.14} \\
			\hline
		\end{tabular}
	}
\end{table*}

\subsection{Results on Unsupervised Image Translation}  
\guo{In this section, we compare the performance of the proposed JWDM with the following baseline methods on unsupervised image translation.}
\begin{itemize}
	\item \textbf{CoGAN \cite{liu2016coupled}}
	\guo{conducts image translation by finding a latent representation that generates images in the source domain and then rendering this latent representation into the target domain.}
	\item \textbf{CycleGAN \cite{zhu2017unpaired}}
	uses an adversarial loss and a cycle-consistent loss to learn a cross-domain mapping for unsupervised image-to-image translation.
	\item \textbf{UNIT \cite{liu2017unsupervised}} learns the joint distribution of images in different domains. It is trained with the images from the marginal distributions in the individual domain.
	\item \textbf{AGGAN \cite{alami2018attention}} introduces unsupervised attention mechanisms that are jointly trained with the generators and discriminators to improve the translation quality.
\end{itemize}

\paragraph{Comparisons with state-of-the-art methods.}
\guo{
	We conduct experiments on five image translation tasks on Cityscapes and SYNTHIA, such as scene2segmentation, segmentation2scene, winter2spring, winter2summer and winter2fall. Quantitative and visual results are shown in Table~\ref{tab:image2image} and Figure~\ref{fig:i2iresults}, respectively.}

\noindent\textbf{Quantitative Comparisons.} From Table~\ref{tab:image2image}, JWDM is able to learn a good joint distribution and consistently outperforms the considered methods on all the image translation tasks. 
On the contrary, CoGAN gets the worst results because it cannot directly translate the input images but samples latent variables to generate the target images.
For CycleGAN and AGGAN, because they cannot exploit the cross-domain information, the translated images may lose some information.
Besides, UNIT is hard to learn a good joint distribution so that the results are worse than our JWDM.

\noindent\textbf{Visual Comparisons.} From Figure~\ref{fig:i2iresults}.
CoGAN translates images with the worst quality and contains noises.
The translated images of CycleGAN and AGGAN are slightly worse and lose some information.
Besides, the translation result of UNIT is better than the above three comparison methods, but its translated image is not sharp enough.
Compared with these methods, JWDM generates more accurate translated images with sharp structure by exploiting sufficient cross-domain information. These results demonstrate the effectiveness of our method in directly learning the joint distribution between different domains.

\begin{figure*}[t]
	\centering
	\includegraphics[width = 2.0\columnwidth]{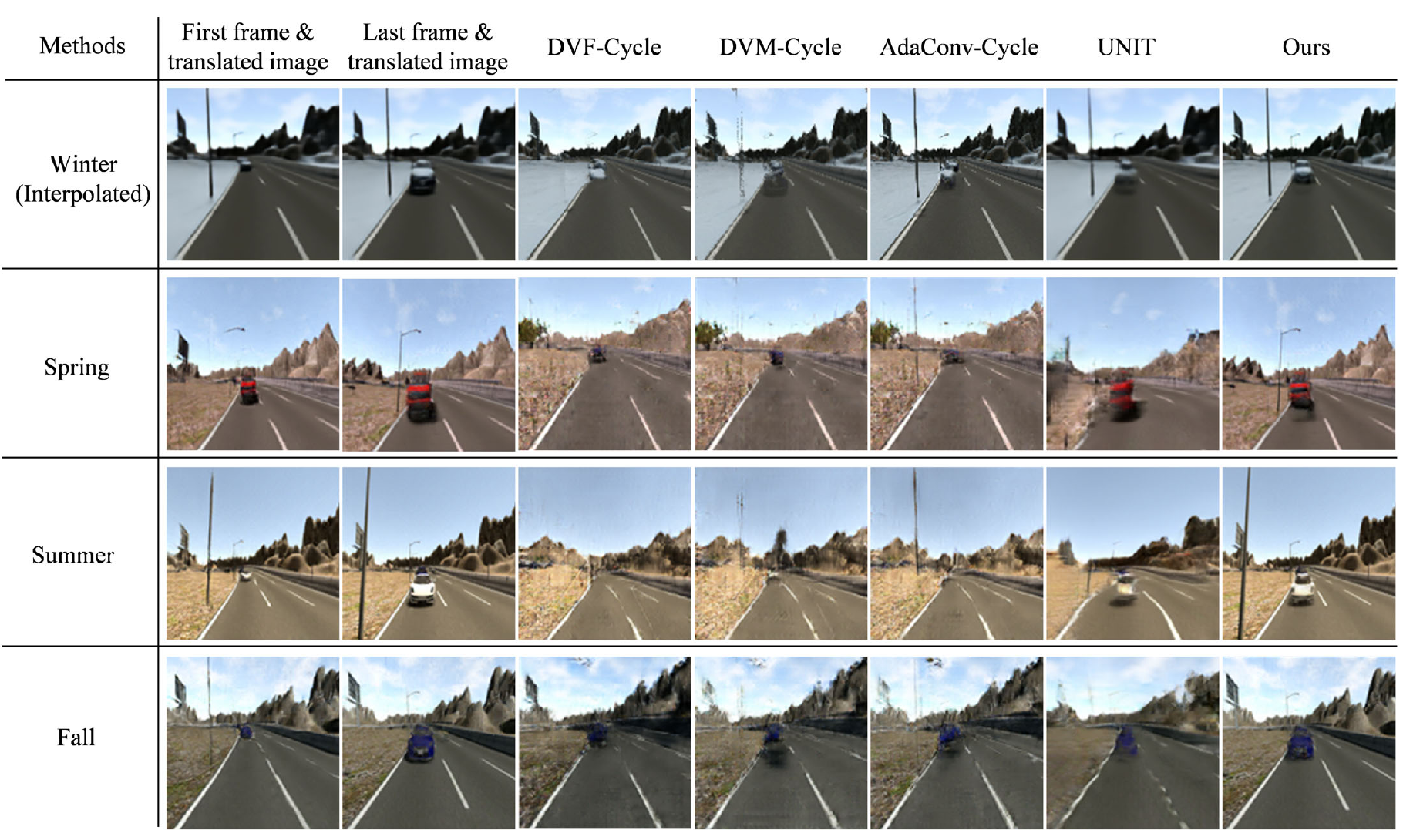}
	\caption{Comparisons of different methods for season translation on SYNTHIA. Top row: the synthesized video in the winter domain. 
		Rows 2-4: The corresponding translated video in the domains of the other three seasons, \emph{i.e.}, spring, summer and fall.
	}
	\label{fig:synthia1}
\end{figure*}

\subsection{Results on Cross-domain Video Synthesis}
\guo{In this experiment, we apply our proposed method to cross-domain video synthesis by performing video frame interpolation and translation in two different domains. 
Specifically, we use the first and ninth video frames in the source domain as input and interpolate the intermediate seven video frames in two domains simultaneously.}

We consider several state-of-the-art baseline methods, including UNIT~\cite{liu2017unsupervised} with latent space interpolation and several constructed variants of CycleGAN~\cite{zhu2017unpaired} with different view synthesis methods. 
For these constructed baselines, we first conduct video frame interpolation and then perform image-to-image translation. 
All considered baselines are shown as follows.

\begin{itemize}
	\item
	\textbf{UNIT \cite{liu2017unsupervised}}
	is an unsupervised image-to-image translation method. 
	Because of the usage of a shared latent space, UNIT is able to conduct interpolations on the latent code of two domains to perform video synthesis.
	\item
	\textbf{DVF-Cycle}
		combines the view synthesis method DVF~\cite{liu2017video} with CycleGAN.
		To be specific, DVF produces videos by video interpolation in one domain, and CycleGAN translates the videos from one domain to another domain.
	\item
	\noindent \textbf{DVM-Cycle}
	uses a geometrical view synthesis DVM \cite{ji2017deep} for video synthesis, and then uses CycleGAN to translate the generated video to another domain. 
	\item
	\textbf{AdaConv-Cycle}
	\guo{combines a state-of-the-art video interpolation method AdaConv~\cite{niklaus2017video} with a pre-trained CycleGAN model. We term it AdaConv-Cycle in the following experiments.}
	\item
	\textbf{Slomo-Cycle}
	\guo{applies the video synthesis method Super Slomo~\cite{jiang2018super} to perform video interpolation in the source domain. Then, we use  pre-trained CycleGAN to translate the synthesized video frames into the target domain. We term it Slomo-Cycle in this paper.}
\end{itemize}

\paragraph{Quantitative Comparisons.}\label{exp:quantitative}
We compare the performance on Cityscapes and SYNTHIA and show the results in Table \ref{tab:cityscapes_synthia}.
For IS, our JWDM achieves comparative performance on scene2segmentation task, while achieves the best performance on other four tasks.
Moreover, JWDM consistently outperforms the baselines in terms of both FID and FID4Video scores.
It means that our method produces frames and videos of promising quality by exploiting the cross-domain correlations.
The above observations demonstrate the superiority of our method over other methods.

\vspace{5pt}
\noindent\textbf{Visual Comparisons.}\label{exp:visual}
\guo{We compare the visual results of different methods on the following three tasks. \footnote{Due to the page limit, more visual results are shown in the supplementary materials.}}

\vspace{5pt}
\noindent\textbf{\emph{(i) Visual results on Cityscapes.}}
We first interpolate videos in the cityscape domain and then translate them to the segmentation domain.
In Figure \ref{fig:cityscapes1}, we compare the visual quality of both the interpolated and the translated images. 
From Figure \ref{fig:cityscapes1} (top), our proposed method is able to produce sharper cityscape images and yields more accurate results in the semantic segmentation domain, which significantly outperforms the baseline methods. Besides, we can drawn the same conclusions in Figure \ref{fig:cityscapes1} (bottom). 

\vspace{5pt}
\noindent\textbf{\emph{(ii) Visual results on SYNTHIA.}}
We further evaluate the performance of our method on SYNTHIA. 
Specifically, we synthesize videos among the domains of four seasons shown in Figure~\ref{fig:synthia1}.
\textbf{First}, our method is able to produce sharper images when interpolating the missing in-between frames (see top row of Figure \ref{fig:synthia1}). 
\textbf{Second}, the translated frames in the spring, summer and fall domains are more photo-realistic than other baseline methods (see the shape of cars in Figure \ref{fig:synthia1}). 
These results demonstrate the efficacy of our method in producing visually promising videos in different domains.

\begin{figure*}[t]
	\centering
	\includegraphics[width = 1.75\columnwidth]{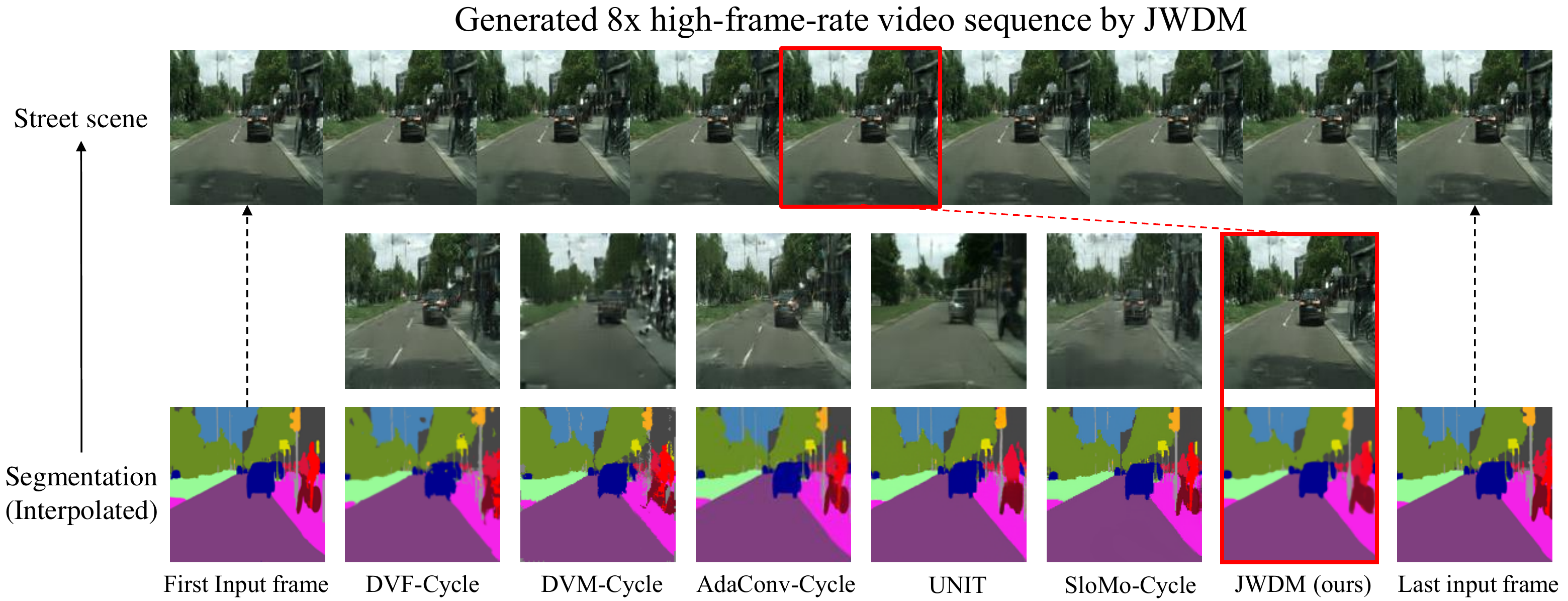}
	\caption{Comparisons of different methods for high-frame-rate image-to-video synthesis, \emph{segmentation2scene}. Top: generated 8$\times$ high-frame-rate video sequence by JWDM. Middle: Interpolated frames of different methods in the target domain. Bottom: Interpolated frames of different methods in the source domain.}
	\label{fig:slowvideo}
\end{figure*}

	\vspace{10pt}
\noindent\textbf{\emph{(iii) High-frame-rate cross-domain video synthesis.}} 
\guo{We investigate the performance of high-frame-rate video synthesis on Cityscapes. Unlike traditional video synthesis studied in the previous section, we use two consecutive video frames in the source domain as input, and interpolate the intermediate 7 frames, \ie, 8$\times$ frame rate up-conversion. 
The synthesized results are shown in Figure~\ref{fig:slowvideo}.} \footnote{Due to the page limit, we only show one synthesized frame of the video for different methods, see supplementary materials for full translated video.}

\guo{In this experiment, we take segmentation frames as the source domain and scene frames as the target domain to perform cross-domain video synthesis. \textbf{First}, we compare the quality of interpolated images in the source domain (see bottom row of Figure~\ref{fig:slowvideo}). It is clear that our method is able to interpolate better frames than most video synthesis methods trained with the ground-truth intermediate frames. \textbf{Second,} we also compare the translated frames in the target domain (see middle row of Figure~\ref{fig:slowvideo}). From Figure~\ref{fig:slowvideo}, our method is able to generate sharper images in the target domain with the help of well learned joint distribution. \textbf{Last}, we show the entire video sequence in the top row of Figure~\ref{fig:slowvideo}. The proposed JWDM is able to produce smooth video sequence in two domains simultaneously. 
}

\subsection{Influence of $\lambda_{z}$ for Adversarial Loss on $\gZ$}\label{exp:influence}
\mo{In this section, we evaluate the influences of the trade-off parameter $\lambda_{z}$ over the adversarial loss on $\gZ$ in Eqn. (\ref{prob:new_obj}).
Specifically, we compare FID and FID4Video with different $\lambda_{z}$ on the winter$ {\leftrightarrow} $summer image translation task.
The value of  $\lambda_{z}$ is selected among [0.01, 0.1, 1, 10].
The quantitative results are shown in Table \ref{tab:influence_rho}.
} 

\mo{From Table \ref{tab:influence_rho}, JWDM achieves the best performance when setting $\lambda_{z}$ to 0.1. 
When increasing the value of $\lambda_{z}$ to 1 and 10, the performance degrades gradually.
The same phenomenon happens when decreasing $\lambda_{z}$ to 0.01.
This means that when setting $\lambda_{z}=0.1$, 
we can achieve better trade-off between the optimization over latent space and data space, 
and thus obtain better performance.}


\begin{table}[t]
	\caption{Influence of $\lambda_{z}$ for the adversarial loss on $\gZ$. We compare the results of winter$\leftrightarrow$summer in terms of FID and FID4Video.}
	\label{tab:influence_rho}
	\centering
	\resizebox{0.4\textwidth}{!}{
		\begin{tabular}{c|c|c|c|c}
			\hline
			\multirow{2}[0]{*}{$\lambda_{z}$} & \multicolumn{2}{c|}{winter2summer} & \multicolumn{2}{c}{summer2winter}  \\
			\cline{2-5}
			& FID & FID4Video & FID & FID4Video \\
			\hline
			0.01 & 94.91 & 20.29 & 107.65 & 18.90 \\
			0.1 & 77.12 & 17.99 & 89.03 & 17.36 \\
			1 & 89.07 & 21.04 & 102.18 & 18.63  \\
			10 & 101.07 & 23.66 & 108.47 & 20.50 \\
			\hline
		\end{tabular}
	}
\end{table}

\section{Conclusion} \label{sec:conclusion}
In this paper, we have proposed a novel joint Wasserstein Distribution Matching (JWDM) method to match joint distributions in different domains.
Relying on optimal transport theory, JWDM is able to exploit cross-domain correlations to improve the performance.
Instead of directly optimizing the primal problem of Wasserstein distance between joint distributions, we derive an important theorem to solve a simple optimization problem.
Extensive experiments on unsupervised image translation and cross-domain video synthesis demonstrate the superiority of the proposed method.


{\small
\balance
\bibliographystyle{ieee}
\bibliography{JWDM}
}

\onecolumn
\vskip 0.3in
\clearpage
\textbf{\Large{~~~~~Supplementary Materials for ``Joint Wasserstein Distribution Matching''}}
\vspace{20pt}

\setcounter{section}{0}

\section{Proof of Theorem \ref{thm:equ_problem}}\label{sec:theory} 
\begin{proof}
	We denote by $ \gP(X {\sim} P_X, {X'} {\sim} P_{G_1}) $ and $ \gP(Y {\sim} P_Y, {Y'} {\sim} P_{G_2}) $ the set of all joint distributions of $ (X, {X'}) $ and $ (Y, {Y'}) $ with marginals $ P_X, P_{G_1} $ and $ P_Y, P_{G_2} $, respectively,
	and denote by $ \gP(P_{\gA}, P_{\gB}) $ the set of all joint distribution of $ P_{\gA} $ and $ P_{\gB} $. Recall the definition of Wasserstein distance $ \gW_c(P_{\gA}, P_{\gB}) $, we have
	\begin{align}
	\gW_c(P_{\gA}, P_{\gB}) 
	=& \inf_{\pi \in \gP(P_{\gA}, P_{\gB})} \E_{(X, Y'; X', Y) \sim \pi} [ c(X, Y'; X', Y)] \label{eqn:AE1}\\
	=& \inf_{\pi \in \gP(P_{\gA}, P_{\gB})} \E_{(X, Y'; X', Y) \sim \pi} [ c_1(X, X') ] 
	+ \inf_{\pi \in \gP(P_{\gA}, P_{\gB})} \E_{(X, Y'; X', Y) \sim \pi} [ c_2(Y', Y) ] \nonumber \\
	=& \inf_{P \in \gP_{X, X'}} \E_{(X, X') \sim P} [c_1(X, X')] 
	+ \inf_{P \in \gP_{Y, Y'}} \E_{(Y', Y) \sim P} [ c_2(Y', Y)] \label{eqn:AE2}\\
	=& \inf_{P \in \gP({P_{X}, P_{G_1}})} \E_{(X, X') \sim P} [c_1(X, X')] 
	+ \inf_{P \in \gP({P_{Y}, P_{G_2}})} \E_{(Y', Y) \sim P} [ c_2(Y', Y)] \label{eqn:AE3}\\
	=& \;\gW_{c_1}(P_{X}, P_{G_1}) +  \gW_{c_2}(P_{G_2}, P_{Y}). \nonumber
	\end{align}
	Line (\ref{eqn:AE1}) holds by the definition of $ \gW_c(P_{\gA}, P_{\gB})  $.
	Line (\ref{eqn:AE2}) uses the fact that the variable pair $ (X, X') $ is independent of the variable pair $ (Y, Y') $, and $ \gP_{X, X'} $ and $ \gP_{Y, Y'} $ are the marginals on $ (X, X') $ and $ Y, Y' $ induced by joint distributions in $ \gP_{X, Y', X', Y} $. 
	In Line (\ref{eqn:AE3}), if $ P_{G_1}(X'|Z) $ and $ P_{G_2}(Y'|Z) $ are Dirac measures (\textit{i.e.}, $ X' {=} G_1(Z) $ and $ Y' {=} G_2(Z) $), we have 
	\[\gP_{X, X'} {=} \gP(P_X, P_{G_1}), \qquad \gP_{Y, Y'} {=} \gP(P_Y, P_{G_2}). \]
	We consider certain sets of joint probability distributions $ \gP_{X, X', Z_1} $ and $ \gP_{Y, Y', Z_2} $ of three random variables $ (X, X', Z_1) \in \gX {\times} \gX {\times} \gZ $ and $ (Y', Y, Z_2) \in \gY {\times} \gY {\times} \gZ $, respectively. 
	We denote by $ \gP(X {\sim} P_X, Z_1 {\sim} P_{Z_1}) $ and $ \gP(Y {\sim} P_Y, Z_2 {\sim} P_{Z_2}) $ the set of all joint distributions of $ (X, Z_1) $ and $ (Y, Z_2) $ with marginals $ P_X, P_{Z_1} $ and $ P_Y, P_{Z_2} $, respectively.
	The set of all joint distributions $ \gP_{X, X', Z_1} $ such that $ X {\sim} P_X, (X', Z_1) {\sim} P_{G_1, Z_1} $ and $ (X' {\Perp} X) | Z_1 $, and likewise for $ \gP_{Y, Y', Z_2} $.
	We denote by $ \gP_{X, X'} $ and $ \gP_{X, Z_1} $ the sets of marginals on $ (X, X') $ and $ (X, Z_1) $ induced by distributions in $ \gP_{X, X', Z_1} $, respectively, and likewise for $ \gP_{Y, Y'} $ and $ \gP_{Y, Z_2} $.
	For the further analyses, we have
	\begin{align}
	&\gW_{c_1}(P_{X}, P_{G_1}) +  \gW_{c_2}(P_{Y}, P_{G_2}) \nonumber\\
	{=}& \inf_{P \in \gP_{X, X', Z_1}} \E_{(X, X', Z_1) \sim P} [ c_1(X, X') ] \label{eqn:ae1}
	{+} \inf_{P \in \gP_{Y', Y, Z_2}} \E_{(Y', Y, Z_2) \sim P} [  c_2(Y', Y) ]  \\
	{=}& \inf_{P \in \gP_{X, X', Z_1}} \E_{P_{Z_1}} \E_{X \sim P(X|Z_1)} \E_{X' \sim P(X'|Z_1)} [ c_1(X, X') ] \label{eqn:ae2} \\
	&{+} \inf_{P \in \gP_{Y', Y, Z_2}} \E_{P_{Z_2}} \E_{Y \sim P(Y|Z_2)} \E_{Y' \sim P(Y'|Z_2)} [ c_2(Y', Y) ] \nonumber \\
	{=}& \inf_{P \in \gP_{X, X', Z_1}} \E_{P_{Z_1}} \E_{X \sim P(X|Z_1)} [ c_1(X, G_1(Z_1)) ] 
	{+} \inf_{P \in \gP_{Y', Y, Z_2}} \E_{P_{Z_2}} \E_{Y \sim P(Y|Z_2)} [ c_2(G_2(Z_2), Y) ] \nonumber \\
	{=}& \inf_{P \in \gP_{X, Z_1}} \E_{(X, Z_1) \sim P} [ c_1(X, G_1(Z_1)) ]  
	{+} \inf_{P \in \gP_{Y, Z_2}} \E_{(Y, Z_2) \sim P} [ c_2(G_2(Z_2), Y) ] \label{eqn:ae4} \\
	{=}& \inf_{P \in \gP(X, Z_1)} \E_{(X, Z_1) {\sim} P} [ c_1(X, G_1(Z_1)) ] 
	{+} \inf_{P \in \gP(Y, Z_2)} \E_{(Y, Z_2) {\sim} P} [ c_2(G_2(Z_2), Y) ] \label{eqn:ae5} \\
	{=}& \inf_{Q \in \gQ_1} \E_{P_{X}} \E_{Q{(Z_1|X)}} [c_1(X, G_1(Z_1))] 
	{+} \inf_{Q \in \gQ_2} \E_{P_{Y}} \E_{Q{(Z_2|Y)}} [c_2(G_2(Z_2), Y)], \label{eqn:ae6} 
	\end{align}
	where $ \gQ_1 {=} \{Q(Z_1|X) | Q_{Z_1} {=} P_Z {=} Q_{Z_2}, P_Y {=}  Q_{Y} \} $ and $ \gQ_2 {=} \{Q(Z_2|Y) | Q_{Z_1} {=} P_Z {=} Q_{Z_2}, P_X {=} Q_{X}\} $ are the set of all probabilistic encoders, where $ Q_{Z_1} $ and $ Q_{Z_2} $ are the marginal distributions of $ Z_1 {\sim} Q(Z_1|X) $ and $ Z_2 {\sim} Q(Z_2|Y) $, where $ X {\sim} P_{X}  $ and $ Y {\sim} P_{Y} $, respectively.  
	
	Line (\ref{eqn:ae1}) uses the tower rule of expectation and Line (\ref{eqn:ae2}) holds by the conditional independence property of $ \gP_{X, X', Z} $. 
	In line (\ref{eqn:ae4}), we take the expectation w.r.t. $ X' $ and $ Y' $, respectively, and use the total probability.
	Line (\ref{eqn:ae5}) follows the fact that $ \gP_{X, Z_1} {=} \gP(X {\sim} P_X, Z_1 {\sim} P_{Z_1}) $ and $ \gP_{Y, Z_2} {=} \gP(Y {\sim} P_Y, Z_2 {\sim} P_{Z_2}) $ since $ \gP(P_X, P_{G_1}), \gP_{X, X', Z_1} $ and $ \gP_{X, Y} $ depend on the choice of conditional distributions $ P_{G_1}(X'|Z_1) $, while $ P_{X, Z_1} $ does not, and likewise for distributions w.r.t. $ Y $ and $ G_2 $. 
	In line (\ref{eqn:ae6}), the generative model $ Q(Z_1|X) $ can be derived from two cases where $ Z_1 $ can be sampled from $ E_1(X) $ and $ E_2(G_2(E_1(X))) $ when $ Q_{Z_1} {=} Q_{Z_2} $ and $ P_Y {=} Q_{Y} $, and likewise for the generative model $ Q(Z_2|Y) $.
\end{proof}

\newpage
\section{Optimization Details}
In this section, we discuss some details of optimization for distribution divergence.
In the training, we use original GAN to measure the divergence $ d(P_X, Q_{X}) $, $ d(P_Y, Q_{Y}) $ and $ d(P_Z, Q_{Z}) $, respectively.

For $ d(P_X, Q_{X}) $, we optimize the following minimax problem:
\begin{align}
&\min_{F} \max_{D_x} \left[ \hat{\E}_{\bx \sim P_X} [\log D_x (\bx)] + \hat{\E}_{\tilde{\bx} \sim Q_X} [\log(1-D_x(\tilde{\bx}))] \right] \\
=&\min_{F} \max_{D_x}\left[ \hat{\E}_{\bx \sim P_X} [\log D_x (\bx)] + \lambda \hat{\E}_{{\bx} \sim P_X} [\log(1-D_x(G_1(E_2(G_2(E_1(\bx)))))))] + (1{-}\lambda) \hat{\E}_{{\by} \sim P_Y} (1{-}D_x(G_1(E_2(\by)))) \right],\nonumber
\end{align}
where $ D_x $ is a discriminator \wrt $ X $, and $ \lambda \in (0, 1) $.

For $ d(P_Y, Q_{Y}) $, we optimize the following minimax problem:
\begin{align}
&\min_{F} \max_{D_y} \left[ \hat{\E}_{\by \sim P_Y} [\log D_y (\by)] + \hat{\E}_{\tilde{\by} \sim Q_Y} [\log(1-D_y(\tilde{\by}))] \right] \\
=&\min_{F} \max_{D_y}\left[ \hat{\E}_{\by \sim P_Y} [\log D_y (\by)] + \lambda \hat{\E}_{{\by} \sim P_Y} [\log(1-D_y(G_2(E_1(G_1(E_2(\by)))))))] + (1{-}\lambda) \hat{\E}_{{\bx} \sim P_X} (1{-}D_y(G_2(E_1(\bx)))) \right],\nonumber
\end{align}
where $D_y$ is a discriminator \wrt $Y$, and $ \lambda \in (0, 1) $.

For $ d(P_Z, Q_{Z}) $, it contains $ d(P_Z, Q_{Z_1}) $ and $ d(P_Z, Q_{Z_2}) $ simultaneously, we optimize the following minimax problem:
\begin{align*}
	d(P_Z, Q_{Z_1}){=}\min_{E_1} \max_{D_z} \left[ \hat{\E}_{\bz \sim P_Z} [\log D_z (\bz)] + \hat{\E}_{{\bx} \sim P_X} [\log(1-D_z(E_1(\bx)))] \right],
\end{align*} 
where $D_z$ is a discriminator \wrt $Z$, and $ \rvz  $ is drawn from the prior distribution $ P_{Z} $.
Similarly, the another term $ d(P_Z, Q_{Z_2}) $ can be written as
\begin{align*}
d(P_Z, Q_{Z_2}){=}\min_{E_2} \max_{D_z} \left[ \hat{\E}_{\bz \sim P_Z} [\log D_z (\bz)] + \hat{\E}_{{\by} \sim P_Y} [\log(1-D_z(E_2(\by)))] \right].
\end{align*}


\newpage
\section{Network Architecture}
The network architectures of JWDM are shown in Tables \ref{table:generator_net} and \ref{table:discriminator_net}. 
We use the following abbreviations: N: the number of output channels, $N_z$: the number of channels for latent variable(set to 256 by default), K: kernel size, S: stride size, P: padding size, IN: instance normalization.

\begin{table}[h]
	\caption{Auto-Encoder architecture.}
	\label{table:generator_net}
	\centering
	\begin{tabular}{c|c|c}
		\hline
		\hline
		\multicolumn{3}{c}{\textbf{Encoder}} \\
		\hline
		\hline 
		{Part} & {Input $ \rightarrow $ Output shape} & {Layer information} \\
		\hline 
		\multirow{3}[0]{*}{Down-sampling}  
		& $ (h, w, 3) \rightarrow (h, w, 64) $ & CONV-(N64, K7x7, S1, P3), IN, ReLU \\	
		& $ (h, w, 64) \rightarrow (\frac{h}{2}, \frac{w}{2}, 128) $ & CONV-(N128, K3x3, S2, P1), IN, ReLU \\
		& $ (\frac{h}{2}, \frac{w}{2}, 128) \rightarrow (\frac{h}{4}, \frac{w}{4}, 256) $ & CONV-(N256, K3x3, S2, P1), IN, ReLU  \\
		\hline 
		\multirow{6}[0]{*}{Bottleneck}  
		& $ (\frac{h}{4}, \frac{w}{4}, 256) \rightarrow (\frac{h}{4}, \frac{w}{4}, 256) $ & Residual Block: CONV-(N256, K3x3, S1, P1), IN, ReLU \\
		& $ (\frac{h}{4}, \frac{w}{4}, 256) \rightarrow (\frac{h}{4}, \frac{w}{4}, 256) $ & Residual Block: CONV-(N256, K3x3, S1, P1), IN, ReLU \\
		& $ (\frac{h}{4}, \frac{w}{4}, 256) \rightarrow (\frac{h}{4}, \frac{w}{4}, 256) $ & Residual Block: CONV-(N256, K3x3, S1, P1), IN, ReLU \\
		& $ (\frac{h}{4}, \frac{w}{4}, 256) \rightarrow (\frac{h}{4}, \frac{w}{4}, 256) $ & Residual Block: CONV-(N256, K3x3, S1, P1), IN, ReLU \\
		& $ (\frac{h}{4}, \frac{w}{4}, 256) \rightarrow (\frac{h}{4}, \frac{w}{4}, 256) $ & Residual Block: CONV-(N256, K3x3, S1, P1), IN, ReLU \\
		& $ (\frac{h}{4}, \frac{w}{4}, 256) \rightarrow (\frac{h}{4}, \frac{w}{4}, 256) $ & Residual Block: CONV-(N256, K3x3, S1, P1), IN, ReLU \\
		\hline
		Embedding Layer & $ (\frac{h}{4}, \frac{w}{4}, 256) \rightarrow (\frac{h}{4}-7, \frac{w}{4}-7, N_z) $ & CONV-($N_z$, K8x8, S1, P0), IN, ReLU \\
		\hline
		\hline
		\multicolumn{3}{c}{\textbf{Decoder}} \\
		\hline
		\hline
		Embedding Layer & $ (\frac{h}{4}-7, \frac{w}{4}-7, N_z) \rightarrow (\frac{h}{4}, \frac{w}{4}, 256) $ & DECONV-(N256, K8x8, S1, P0), IN, ReLU \\
		\hline
		\multirow{6}[0]{*}{Bottleneck}  
		& $ (\frac{h}{4}, \frac{w}{4}, 256) \rightarrow (\frac{h}{4}, \frac{w}{4}, 256) $ & Residual Block: CONV-(N256, K3x3, S1, P1), IN, ReLU \\
		& $ (\frac{h}{4}, \frac{w}{4}, 256) \rightarrow (\frac{h}{4}, \frac{w}{4}, 256) $ & Residual Block: CONV-(N256, K3x3, S1, P1), IN, ReLU \\
		& $ (\frac{h}{4}, \frac{w}{4}, 256) \rightarrow (\frac{h}{4}, \frac{w}{4}, 256) $ & Residual Block: CONV-(N256, K3x3, S1, P1), IN, ReLU \\
		& $ (\frac{h}{4}, \frac{w}{4}, 256) \rightarrow (\frac{h}{4}, \frac{w}{4}, 256) $ & Residual Block: CONV-(N256, K3x3, S1, P1), IN, ReLU \\
		& $ (\frac{h}{4}, \frac{w}{4}, 256) \rightarrow (\frac{h}{4}, \frac{w}{4}, 256) $ & Residual Block: CONV-(N256, K3x3, S1, P1), IN, ReLU \\
		& $ (\frac{h}{4}, \frac{w}{4}, 256) \rightarrow (\frac{h}{4}, \frac{w}{4}, 256) $ & Residual Block: CONV-(N256, K3x3, S1, P1), IN, ReLU \\
		\hline
		\multirow{3}[0]{*}{Up-sampling}
		& $ (\frac{h}{4}, \frac{w}{4}, 256) \rightarrow (\frac{h}{2}, \frac{w}{2}, 128) $ & DECONV-(N128, K3x3, S2, P1), IN, ReLU \\
		& $ (\frac{h}{2}, \frac{w}{2}, 128) \rightarrow (h, w, 64) $ & DECONV-(N64, K3x3, S2, P1), IN, ReLU \\
		& $ (h, w, 64) \rightarrow (h, w, 3) $ & CONV-(N3, K7x7, S1, P3), Tanh \\
		\hline
		\hline
	\end{tabular}
\end{table}

\begin{table}[h]
	\caption{Discriminator network architecture.}
	\label{table:discriminator_net}
	\centering
	\begin{tabular}{c|c|c}
		\hline 
		\hline
		{Layer} & {Input $ \rightarrow $ Output shape} & {Layer information} \\
		\hline 
		Input Layer  & $ (h, w, 3) \rightarrow (\frac{h}{2}, \frac{w}{2}, 64) $ & CONV-(N64, K4x4, S2, P1), Leaky ReLU \\
		\hline
		Hidden Layer & $ (\frac{h}{2}, \frac{w}{2}, 64) \rightarrow (\frac{h}{4}, \frac{w}{4}, 128) $ & CONV-(N128, K4x4, S2, P1), Leaky ReLU \\
		Hidden Layer & $ (\frac{h}{4}, \frac{w}{4}, 128) \rightarrow (\frac{h}{8}, \frac{w}{8}, 256) $ &CONV-(N256, K4x4, S2, P1), Leaky ReLU \\
		Hidden Layer & $ (\frac{h}{8}, \frac{w}{8}, 256) \rightarrow (\frac{h}{16}, \frac{w}{16}, 512) $ &CONV-(N512, K4x4, S2, P1), Leaky ReLU \\
		Hidden Layer & $ (\frac{h}{16}, \frac{w}{16}, 512) \rightarrow (\frac{h}{32}, \frac{w}{32}, 1024) $ &CONV-(N1024, K4x4, S2, P1), Leaky ReLU \\
		Hidden Layer & $ (\frac{h}{32}, \frac{w}{32}, 1024) \rightarrow (\frac{h}{64}, \frac{w}{64}, 2048) $ &CONV-(N2048, K4x4, S2, P1), Leaky ReLU  \\
		\hline
		Output layer & $ (\frac{h}{64}, \frac{w}{64}, 2048) \rightarrow (\frac{h}{64}, \frac{w}{64}, 1) $ & CONV-(N1, K3x3, S1, P1) \\
		\hline
		\hline
	\end{tabular}
\end{table}

\newpage
\section{More Results}

\subsection{Results of Unsupervised Image-to-image Translation}
\begin{figure}[ht]
	\centering
	\includegraphics[width = 1\columnwidth]{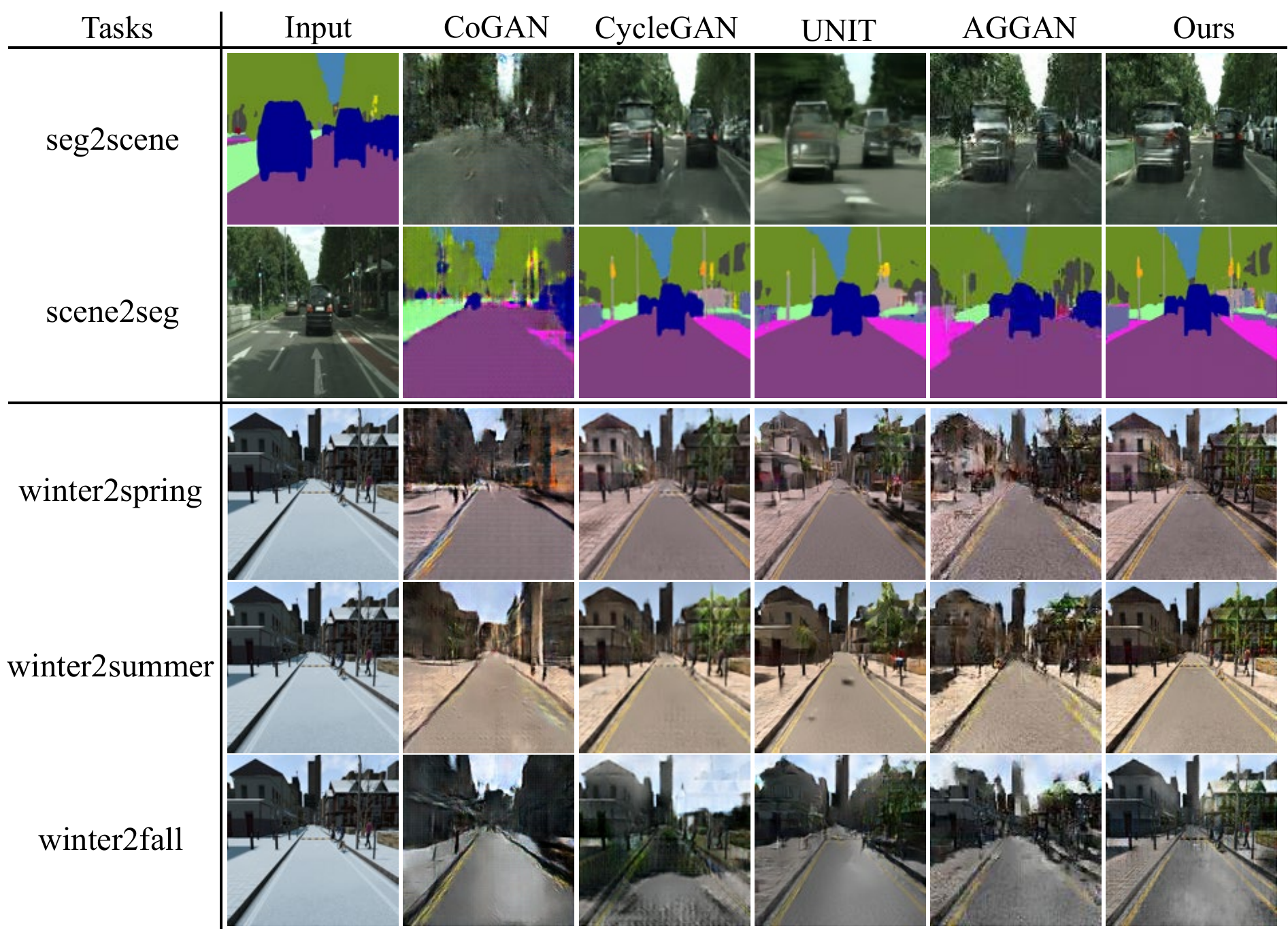}
	\caption{Comparisons with different methods for unsupervised image-to-image translation on Cityscape and SYNTHIA. }
	\label{fig:i2i_one}
\end{figure}

\begin{figure}[t]
	\centering
	\includegraphics[width = 1\columnwidth]{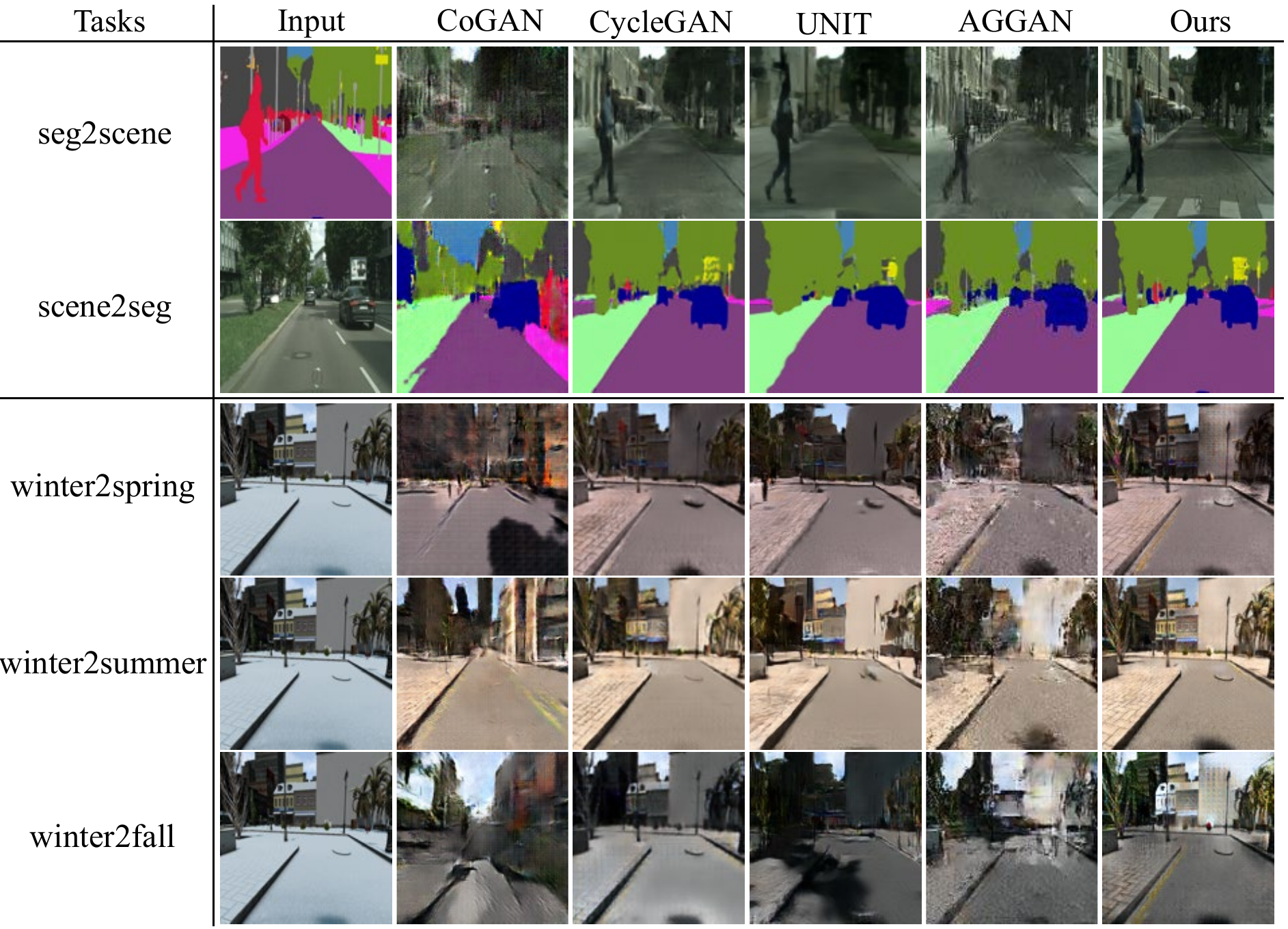}
	\caption{Comparisons with different methods for unsupervised image-to-image translation on Cityscape and SYNTHIA. }
	\label{fig:i2i_two}
\end{figure}

In this section, we provide more visual results for unsupervised image-to-image translation on SYNTHIA and Cityscapes datasets.
The results are shown in Figure ~\ref{fig:i2i_one} and Figure ~\ref{fig:i2i_two}.

Compared with the baseline methods, JWDM generates more accurate translated images with sharp structure by exploiting sufficient cross-domain information. These results demonstrate the effectiveness of our method in directly learning the joint distribution between different domains.

\clearpage
\subsection{Results of Unsupervised Cross-domain Video Synthesis}

In this section, we provide more visual results for cross-domain video synthesis on SYNTHIA and Cityscapes datasets. 
Note that we use the first and eighth video frames in the source domain as input and interpolate the intermediate seven video frames in two domains simultaneously.

\begin{figure}[ht]
	\centering	\includegraphics[width = 1\columnwidth]{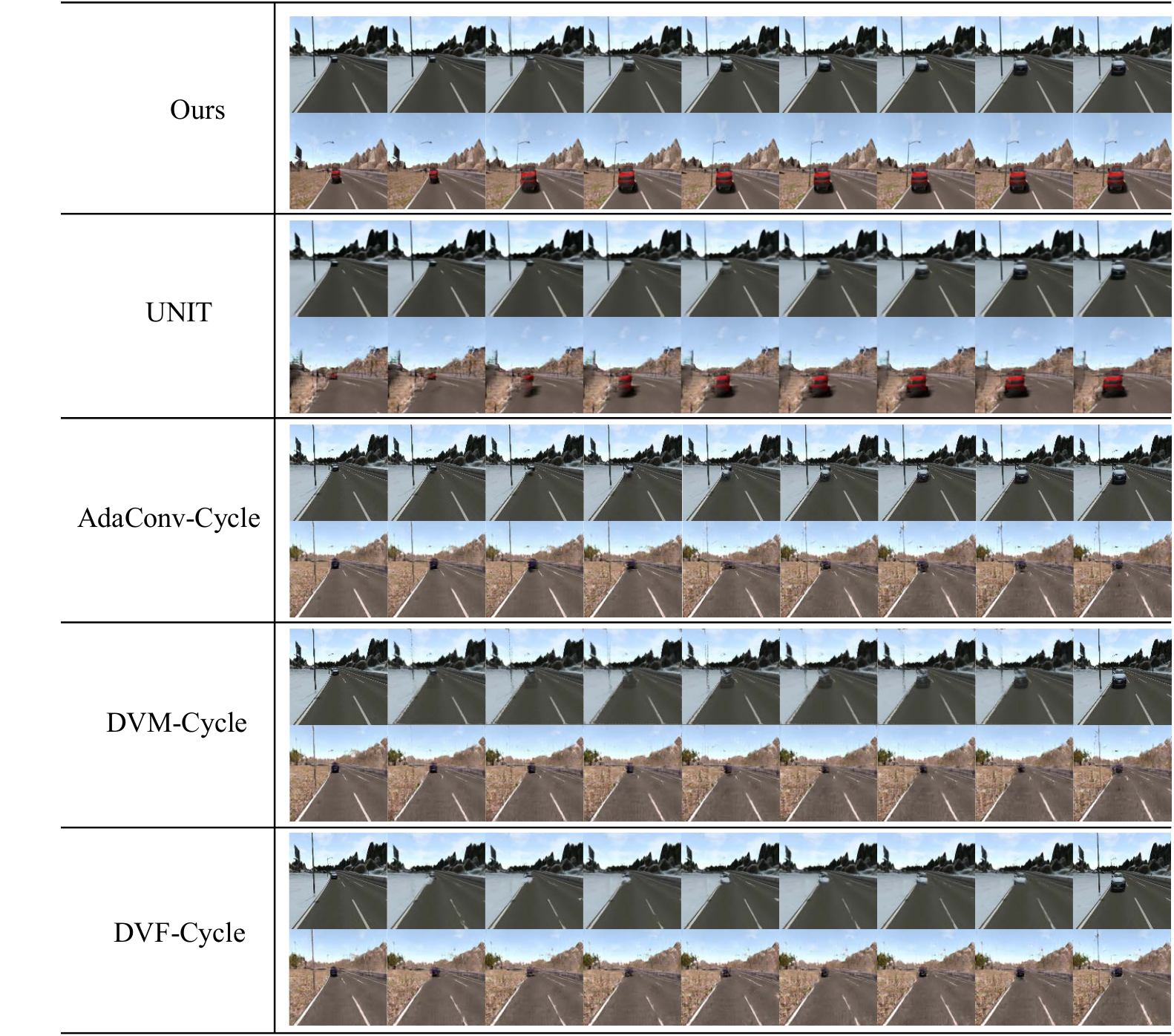}
	\caption{Comparisons of different methods for season winter$\rightarrow$spring translation on SYNTHIA dataset. The figure shows all frames of a video synthesized and translated by these mehtods. }
	\label{fig:synthia_w2sp}
\end{figure}

\begin{figure}[t]
	\centering
	\includegraphics[width = 1\columnwidth]{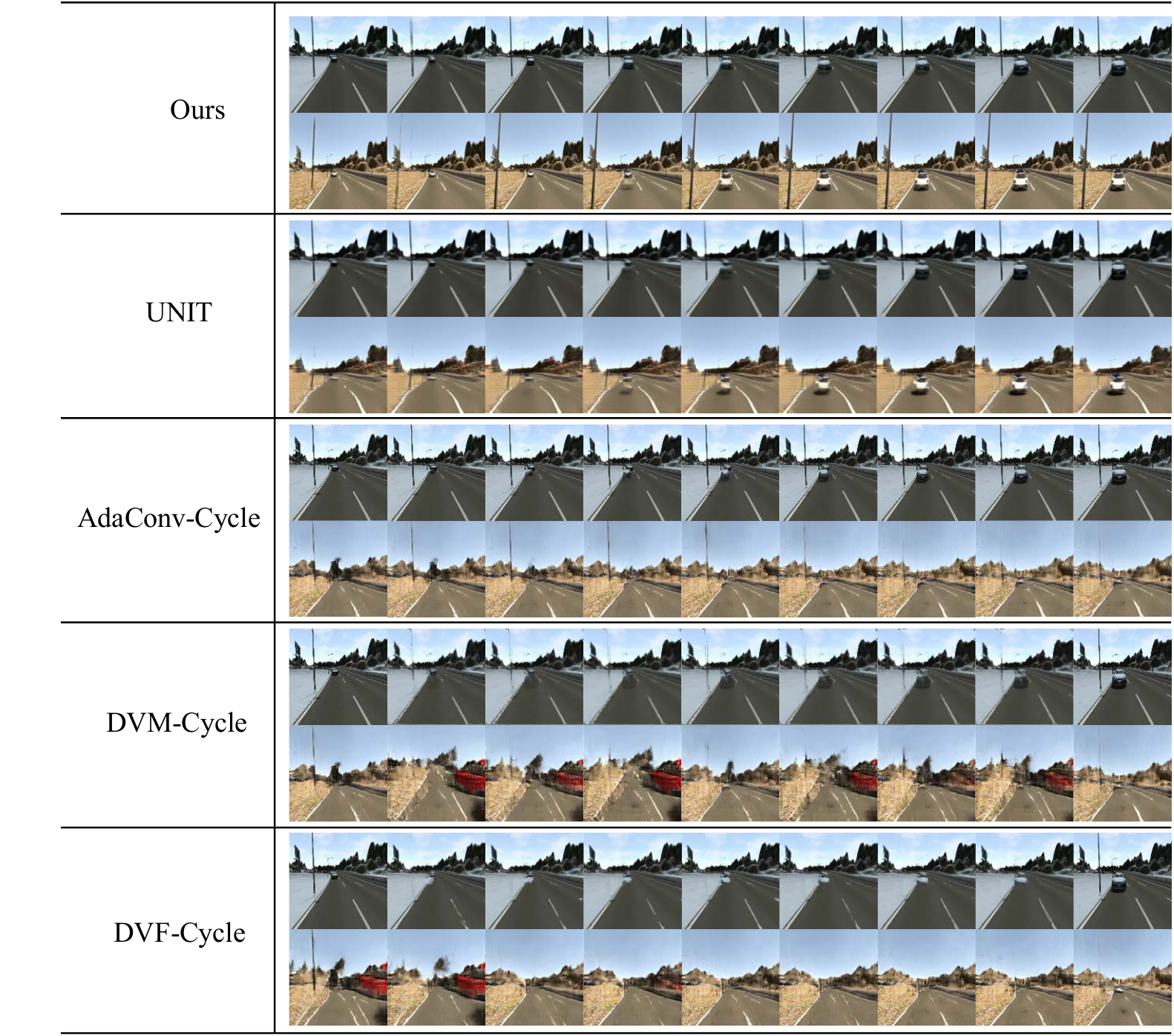}
	\caption{Comparisons of different methods for season winter$\rightarrow$summer translation on SYNTHIA dataset. The figure shows all frames of a video synthesized and translated by these mehtods. }
	\label{fig:synthia_w2su}
\end{figure}

\begin{figure}[t]
	\centering
	\includegraphics[width = 1\columnwidth]{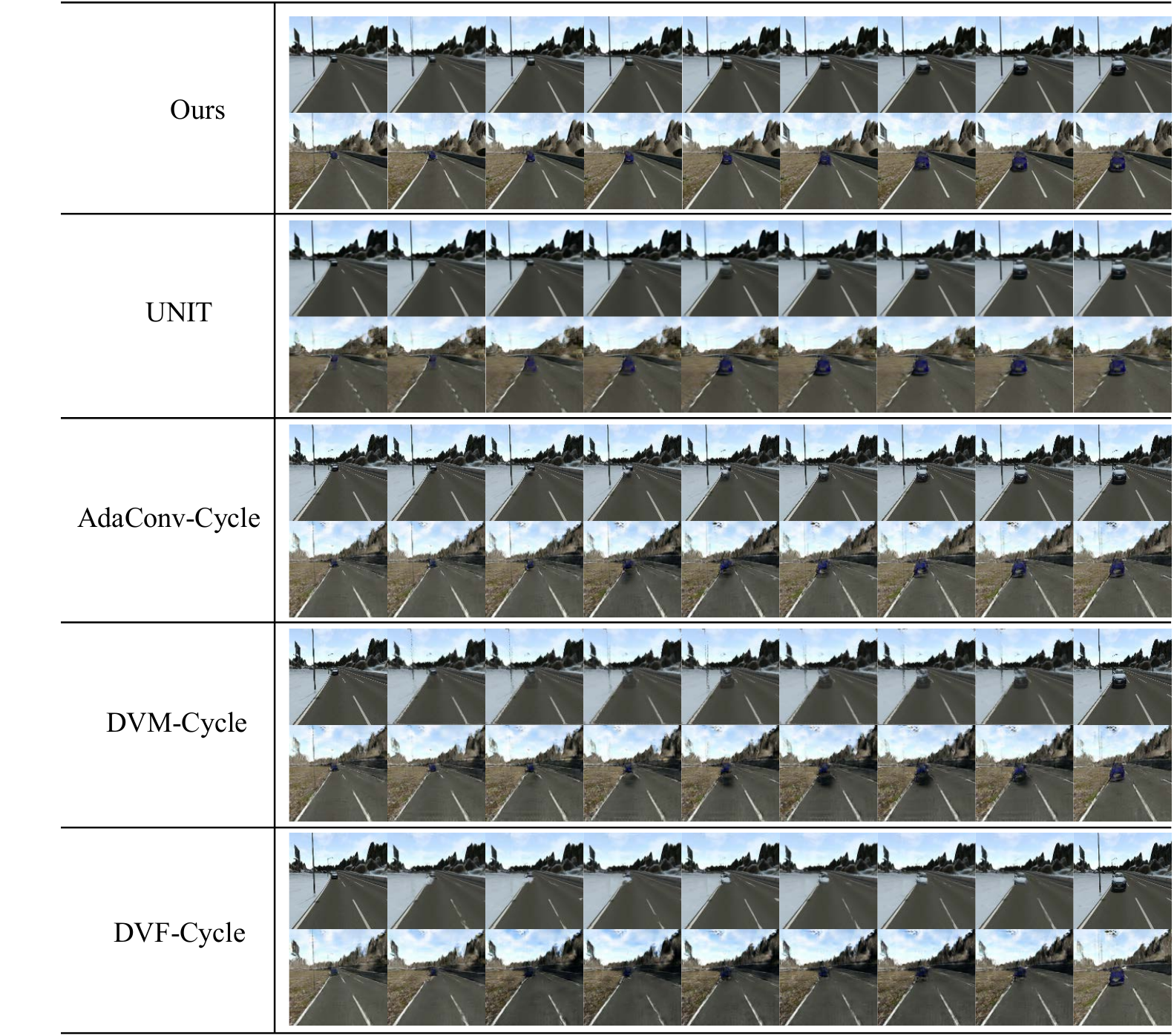}
	\caption{Comparisons of different methods for season winter$\rightarrow$fall translation on SYNTHIA dataset. The figure shows all frames of a video synthesized and translated by these mehtods. }
	\label{fig:synthia_w2f}
\end{figure}

\begin{figure}[t]
	\centering
	\includegraphics[width = 1\columnwidth]{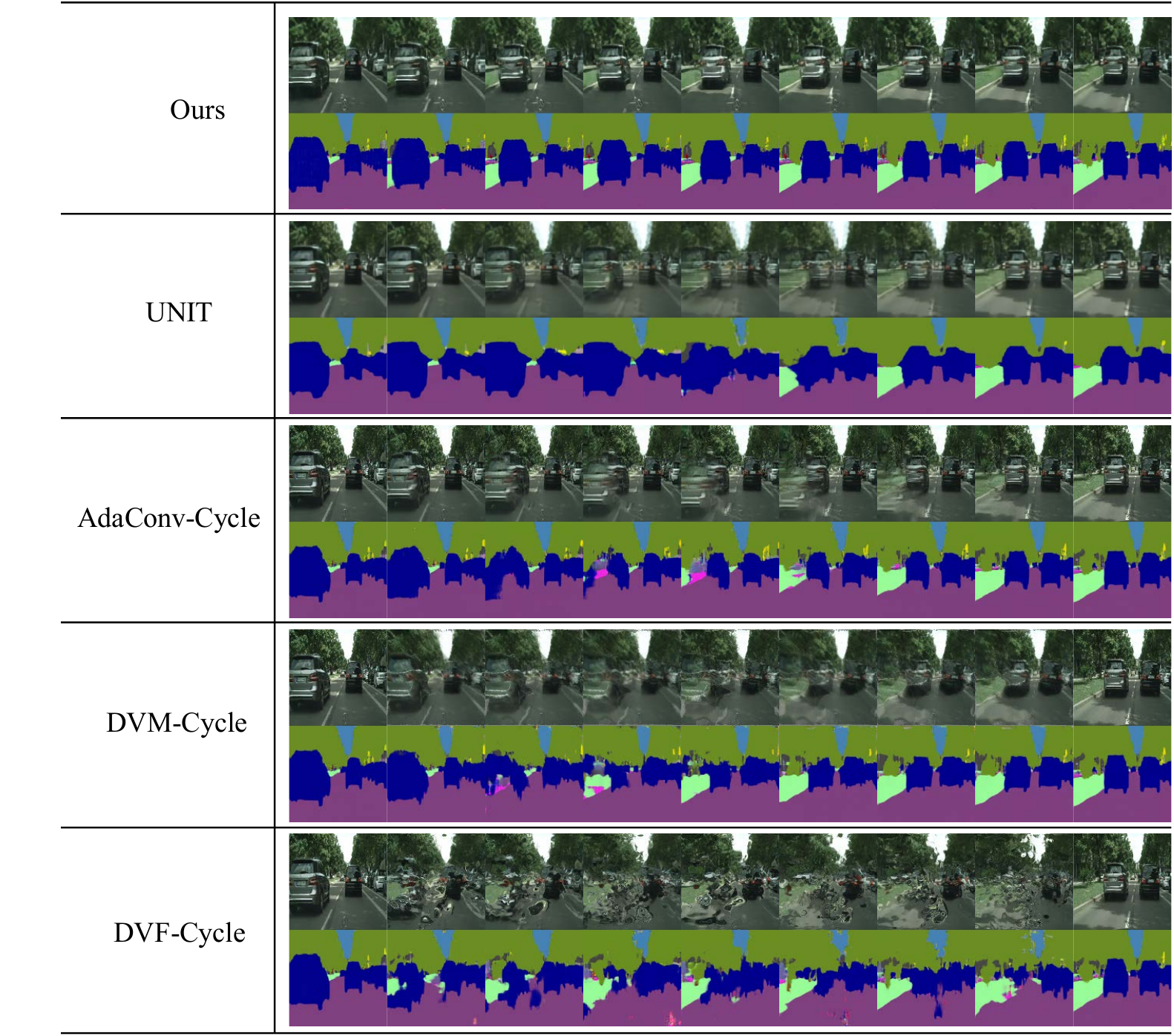}
	\caption{Comparisons of different methods for photo$\rightarrow$segmentation translation on Cityscapes dataset. The figure shows all frames of a video synthesized and translated by these mehtods.}
	\label{fig:cityscapes_c2s1}
\end{figure}

\begin{figure}[t]
	\centering
	\includegraphics[width = 1\columnwidth]{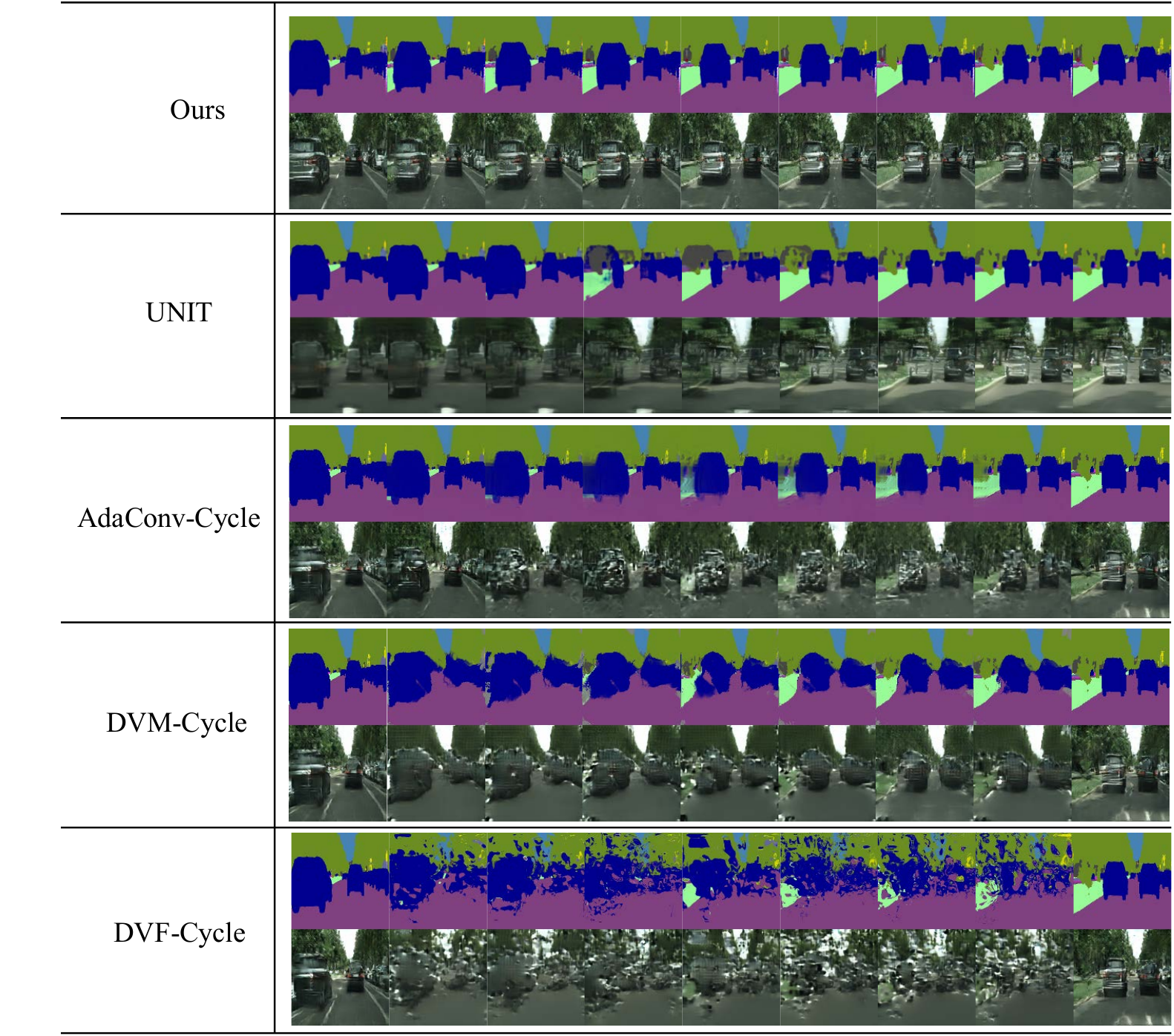}
	\caption{Comparisons of different methods for segmentation$\rightarrow$photo translation on Cityscapes dataset. The figure shows all frames of a video synthesized and translated by these mehtods.}
	\label{fig:cityscapes_s2c1}
\end{figure}

\begin{figure}[t]
	\centering
	\includegraphics[width = 1\columnwidth]{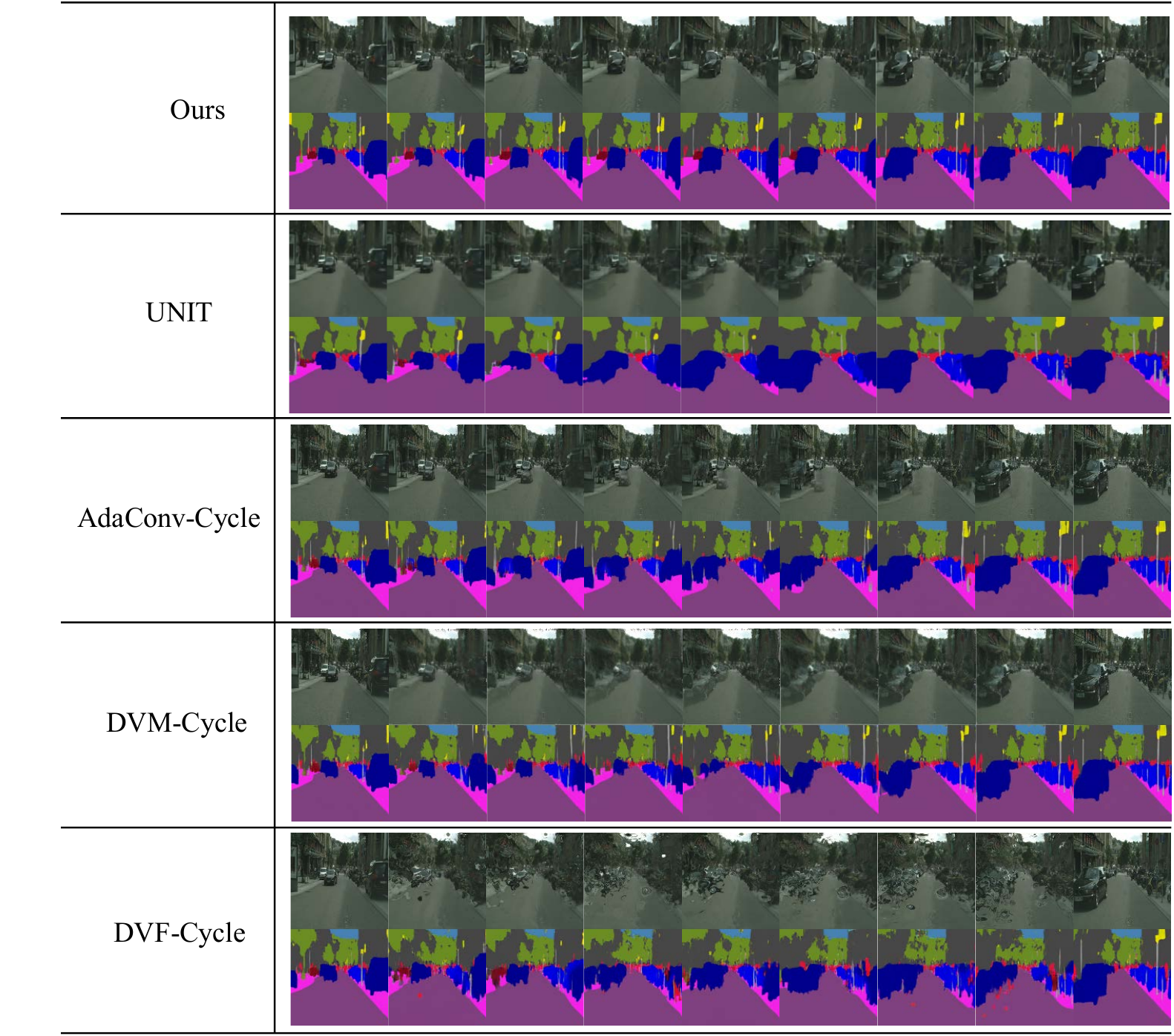}
	\caption{Comparisons of different methods for photo$\rightarrow$segmentation translation on Cityscapes dataset. The figure shows all frames of another video synthesized and translated by these mehtods.}
	\label{fig:cityscapes_c2s2}
\end{figure}

\begin{figure}[t]
	\centering
	\includegraphics[width = 1\columnwidth]{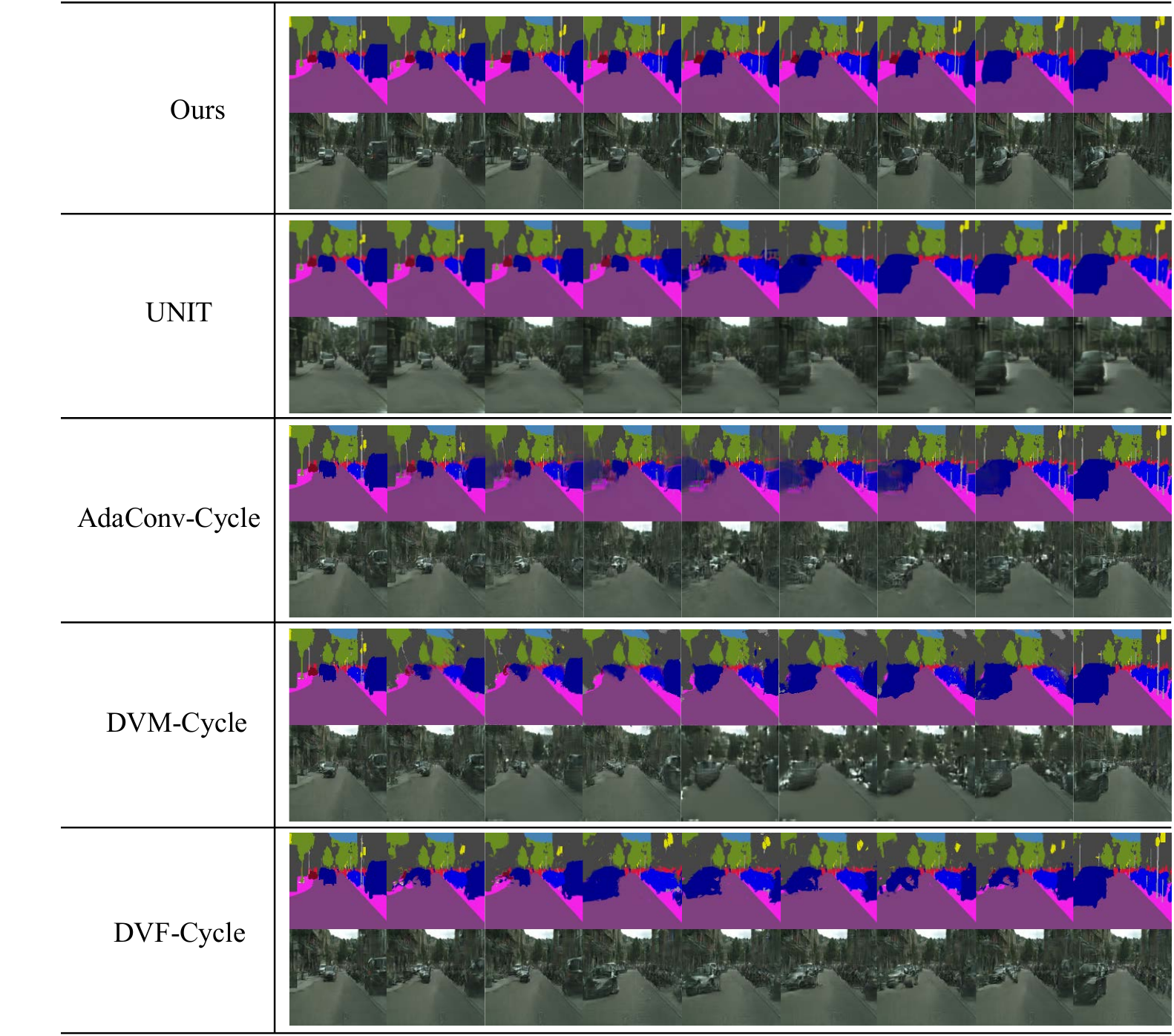}
	\caption{Comparisons of different methods for segmentation$\rightarrow$photo translation on Cityscapes dataset. The figure shows all frames of another video synthesized and translated by these mehtods.}
	\label{fig:cityscapes_s2c2}
\end{figure}

Figure ~\ref{fig:synthia_w2sp}, Figure ~\ref{fig:synthia_w2su} and Figure ~\ref{fig:synthia_w2f} are video synthesis results on winter $\rightarrow$ \{spring, summer and fall\}.
Figure ~\ref{fig:cityscapes_c2s1}, Figure ~\ref{fig:cityscapes_s2c1} and Figure ~\ref{fig:cityscapes_c2s2}, Figure ~\ref{fig:cityscapes_s2c2} are two sets of video synthesis results on photo$\leftrightarrow$segmentation respectively. Evidently, both visual results on SYNTHIA and Cityscapes datasets show that our method produces frames and videos of promising quality and consistently outperforms all the baselines. 

\vspace{20pt}


\end{document}